\documentclass[10pt,journal,compsoc]{IEEEtran}

\ifCLASSOPTIONcompsoc
  \usepackage[noadjust]{cite}
\else
  \usepackage{cite}
\fi

\ifCLASSINFOpdf
   \usepackage[pdftex]{graphicx}
   \graphicspath{{img/}}
\else
\fi

\usepackage{subcaption}

\usepackage{amsmath}
\DeclareMathOperator*{\argmax}{argmax} 
\usepackage{algorithmic}
\usepackage{array}
\usepackage{url}
\usepackage{multirow, booktabs}
\usepackage{enumitem}
\usepackage{amsmath,amsfonts,amssymb}
\usepackage{pgfplotstable}
\pgfplotsset{compat=1.12}
\usetikzlibrary{patterns}
\usepgfplotslibrary{groupplots}

\usepackage{soul}
\usepackage{makecell} 
\usepackage{hyperref}
\usepackage{rotating}

\begin{document}
\title{TransAlign: Fully Automatic and Effective Entity Alignment for Knowledge Graphs}
\author{Rui~Zhang*,
        Xiaoyan~Zhao,
        Bayu~Distiawan~Trisedya,
        Min~Yang,
        Hong~Cheng,
        and~Jianzhong~Qi
\IEEEcompsocitemizethanks{
    \IEEEcompsocthanksitem R. Zhang is with Tsinghua University. \\
    E-mail: \{rayteam\}@yeah.net. *R.~Zhang is the corresponding author.
    \IEEEcompsocthanksitem X. Zhao, and H. Cheng are with The Chinese University of Hong Kong, Hong Kong, China. \\
    E-mail: \{xzhao, hcheng\}@se.cuhk.edu.hk
    \IEEEcompsocthanksitem B.D. Trisedya is with Universitas Indonesia.\protect\\
	E-mail: \{b.distiawan\}@cs.ui.ac.id
    \IEEEcompsocthanksitem M. Yang is with the Shenzhen Institute of Advanced Technology, Chinese Academy of Sciences, Shenzhen, China.\protect\\
	E-mail: \{min.yang\}@siat.ac.cn
	\IEEEcompsocthanksitem J. Qi is with The University of Melbourne.\protect\\
	E-mail: \{jianzhong.qi\}@unimelb.edu.au
}
\thanks{Manuscript received XXXXXX XX, XXXX; revised XXXXXX XX, XXXX.}}

\markboth{IEEE Transactions on Knowledge and Data Engineering}%
{TransAlign: Fully Automatic and Effective Entity Alignment for Knowledge Graphs}

\IEEEtitleabstractindextext{%
 
\begin{abstract}\label{abstract}
	The task of entity alignment between knowledge graphs (KGs) aims to identify every pair of entities from two different KGs that represent the same entity. 
	Many machine learning-based methods have been proposed for this task. However, to our best knowledge, existing methods all require \emph{manually crafted} seed alignments, which are expensive to obtain. 
	In this paper, we propose the first fully automatic alignment method named TransAlign, which does not require any manually crafted seed alignments. 
	Specifically, for predicate embeddings, TransAlign constructs a predicate-proximity-graph to automatically capture the similarity between predicates across two KGs by learning the attention of entity types. For entity embeddings, TransAlign first computes the entity embeddings of each KG independently using TransE, and then shifts the two KGs' entity embeddings into the same vector space by computing the similarity between entities based on their attributes. Thus, both predicate alignment and entity alignment can be done without manually crafted seed alignments. TransAlign is not only fully automatic, but also highly effective.
	Experiments using real-world KGs show that TransAlign improves the accuracy of entity alignment significantly compared to state-of-the-art methods. 
\end{abstract}
}

\maketitle
\IEEEdisplaynontitleabstractindextext
\IEEEpeerreviewmaketitle

\IEEEraisesectionheading{
\section{Introduction} 
	\label{kba-intro}}
Knowledge bases in the form of \emph{knowledge graphs} (KGs) have been used in many applications, including question answering systems~\cite{wu2017image,wang2017visual}, dialogue systems~\cite{xu2019end}, and recommender systems ~\cite{zhang2016rec}. Many KGs have been created separately for particular purposes. The same real-world entity may exist in different forms in different KGs. For example, a village named \texttt{Kromsdorf} in Germany is a real-world entity that exists in two different KGs, LinkedGeoData \cite{stadler2012linkedgeodata} and DBpedia \cite{auer2007dbpedia}. This entity is denoted in the form of \texttt{lgd:240111203} in LinkedGeoData but in the form of \texttt{dbp:Kromsdorf} in DBpedia. Usually, these KGs complement each other in terms of the number of entities each KG contains, and the types of information related to each entity. Therefore, we may merge two KGs into one with more entities and richer information related to each entity. To merge two KGs, we need to solve the problem of \emph{entity alignment}, which is to identify every pair of entities from the two KGs that correspond to the same real-world entity. Existing methods require significant manual work (e.g., manually crafted seed alignments), and the accuracy of the alignment is low. In this paper, we propose a novel method to this problem, which is fully automatic and effective (i.e., the alignment result is of high accuracy).

\begin{table}[t!]
	\centering
	\caption{Knowledge graph alignment example.}
	\begin{small}
		\begin{tabular}{@{}l@{}}
			\toprule[2pt]
			\midrule
			\multicolumn{1}{c}{$\mathcal{G}_1$} \\
			\midrule
			$\langle$\texttt{lgd:240111203,lgd:population,1595}$\rangle$ \\
			$\langle$\texttt{lgd:240111203,rdfs:label,"Kromsdorf"}$\rangle$ \\
			$\langle$\texttt{lgd:240111203,geo:lat,50.9988888889}$\rangle$ \\
			$\langle$\texttt{lgd:240111203,lgd:alderman,"B. Grobe"}$\rangle$ \\
			$\langle$\texttt{lgd:240111203,lgd:is\_in,lgd:51477}$\rangle$ \\
			\midrule
			\multicolumn{1}{c}{$\mathcal{G}_2$} \\
			\midrule
			$\langle$\texttt{dbp:Kromsdorf,rdfs:label,"Kromsdorf"}$\rangle$ \\
			$\langle$\texttt{dbp:Kromsdorf,geo:lat,50.9989}$\rangle$ \\
			$\langle$\texttt{dbp:Kromsdorf,dbp:populationTotal,1595}$\rangle$ \\
			$\langle$\texttt{dbp:Kromsdorf,dbp:located\_in,dbp:Germany}$\rangle$ \\
			$\langle$\texttt{dbp:Kromsdorf,dbp:district,dbp:Weimarer}$\rangle$ \\
			\midrule
			\multicolumn{1}{c}{Merged $\mathcal{G}_{U}$} \\
			\midrule
			$\langle$\texttt{lgd:240111203,:population,1595}$\rangle$ \\
			$\langle$\texttt{lgd:240111203,:label,"Kromsdorf"}$\rangle$ \\
			$\langle$\texttt{lgd:240111203,:lat,50.9988888889}$\rangle$ \\
			$\langle$\texttt{lgd:240111203,:alderman,"B. Grobe"}$\rangle$ \\
			$\langle$\texttt{lgd:240111203,:is\_in,lgd:51477}$\rangle$ \\
			$\langle$\texttt{lgd:240111203,:district,dbp:Weimarer}$\rangle$ \\
			\midrule
			\bottomrule[2pt]
		\end{tabular}%
	\end{small}
	\label{table-kba-example}%
\end{table}%

We use an example as shown in Table~\ref{table-kba-example} to illustrate the entity alignment problem in detail. Typically, knowledge or real-world facts in KGs are stored in the form of triples, 
and a triple consists of three elements in the form of  $\langle \textit{head}, \textit{predicate}, \textit{tail} \rangle$, where \textit{head} denotes an entity and \textit{tail} denotes either another entity or a literal (attribute value) of the head entity. Here, if \textit{tail} is an entity, the triple is called a \emph{relation triple} and the predicate is called \emph{relation predicate}; if \textit{tail} is a literal, the triple is called an \emph{attribute triple} and the predicate is called \emph{attribute predicate}. Table~\ref{table-kba-example} gives an example of two subsets of triples from two KGs, denoted by $\mathcal{G}_1$ and $\mathcal{G}_2$ (we use prefixes \texttt{lgd:} and \texttt{dbp:} to simplify the original spell out). The head entities in these two subsets refer to the same entity \texttt{Kromsdorf}, even though they are in different forms, \texttt{lgd:240111203} and \texttt{dbp:Kromsdorf}. We aim to identify such entities and give them a unified ID such that both KGs can be merged together through them. In Table~\ref{table-kba-example}, $\mathcal{G}_{U}$ denotes the merged KG, where \texttt{lgd:240111203} is used as the unified ID for the entity \texttt{Kromsdorf} which has a set of properties that is the union of the sets of properties from both KGs.

\begin{figure}[t!]
	\begin{center}
		\includegraphics[width=.45\textwidth]{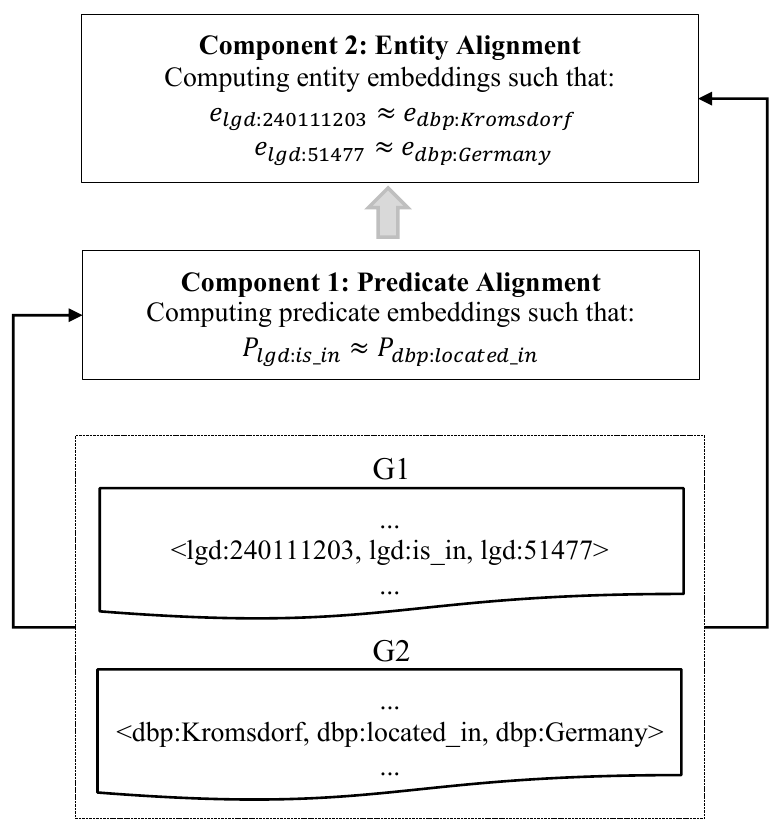}
	\end{center}
	\caption{\label{fig-kba-components} Components of knowledge graph alignment: 
\emph{predicate alignment} and \emph{entity alignment}. }
	\vspace{-3mm}
\end{figure}%
Recently, embedding-based methods have been proposed for KG alignment. 
As illustrated in Fig.~\ref{fig-kba-components}, there are two critical components, 
\emph{predicate alignment} and \emph{entity alignment}:
(i) the embeddings of predicates that represent the same relationship in the two KGs should have similar embeddings in the aligned vector space, e.g., \texttt{lgd:is\_in} and \texttt{dbp:located\_in} should have close embeddings, and (ii) if entity $e_{g_1}$ from $\mathcal{G}_1$ corresponds to the same real-world entity as entity $e_{g_2}$ from $\mathcal{G}_2$, then $e_{g_1}$ should have similar embeddings to that of $e_{g_2}$ in the aligned vector space, e.g., \texttt{lgd:240111203} and \texttt{dbp:Kromsdorf} in Fig.~\ref{fig-kba-components}  should have close embeddings. 

There are mainly two paradigms of KG
embedding for entity alignment, \emph{translation-based methods} and
\emph{Graph Neural Network (GNN-based) methods}. See \cite{zhang2022benchmark} for a comprehensive survey.
Translation-based entity alignment methods \cite{chen2017multilingual,chen2017multigraph,zhu2017iterative} first learn an embedding space for each KG separately, then learn a \emph{transition matrix} to map the embedding space from one KG to the other. The mapping relies on large numbers of seed alignments (i.e., a set of manually crafted aligned triples from the two KGs) to compute the transition matrix. Such manually crafted seed alignments are expensive and error-prone. The other paradigm, GNN-based methods \cite{cao2019multi,sun2020knowledge,xu2019cross,wang2018cross,wu2019relation,ye2019a}, aggregates information from the neighborhood of entities together with the graph structure to compute entity embeddings. Then they align the space of the two KGs via manually crafted seed alignments, which is similar to translation-based methods. 
Moreover, all the existing studies have focused on entity alignment, whereas for predicate alignment, they also rely on manually crafted seeds.
In summary, to our best knowledge, existing methods of both paradigms rely on manually crafted seed alignments.

To address the above problems, we propose a novel method for KG alignment that is not only \emph{fully automatic} (i.e., not involving any manual seed alignments) but also significantly more accurate in aligning entities and predicates (i.e., more effective). 
TransAlign first constructs a predicate-proximity-graph to automatically capture the similarity between predicates across two KGs by learning the attentions of entity types. Then TransAlign computes the entity embeddings of each KG independently using TransE, and shifts the two KGs’ entity embeddings into the same vector space by computing the similarity between entities based on their attributes.
We name our method \emph{TransAlign} as it is a translation-based alignment method. 

Specifically, to achieve predicate embedding alignment without manually crafted seed alignments, we propose a predicate-proximity-graph for approximately computing the predicate embeddings, including both relation predicates and attribute predicates, where each predicate is a vertex that represents a relationship between entity types or literal types (instead of entities or literals). We create such a graph by replacing the head entity and tail entity of KG triples by their corresponding types, which are provided as \texttt{rdfs:type} relationship in the knowledge base. For example, we replace the triples $\langle$\texttt{dbp:Kromsdorf}, \texttt{dbp:located\_in}, \texttt{dbp:Germany}$\rangle$ and $\langle$\texttt{lgd:240111203}, \texttt{lgd:is\_in}, \texttt{lgd:51477}$\rangle$ with the triples $\langle$\texttt{village}, \texttt{dbp:located\_in}, \texttt{country}$\rangle$ and $\langle$\texttt{village}, \texttt{lgd:is\_in}, \texttt{country}$\rangle$, respectively.
Using the predicate-proximity-graph, TransAlign can learn the similarity between predicates from two KGs that represent the same relationships, e.g., the predicates \texttt{dbp:located\_in} and \texttt{lgd:is\_in}. 
Capturing predicate similarity in different KGs via a predicate-proximity-graph has a few challenges. First, each entity often has multiple types. For example, the entity \texttt{Germany} may have multiple entity types \{\texttt{thing, place, location, country}\} in a KG. Second, different KGs may have different schemes of entity types, e.g., in another KG, the entity \texttt{Germany} may have entity types \{\texttt{place, country}\}. Hence, in the predicate-proximity-graph, the head entity and the tail entity may be replaced by multiple entity types. To capture the predicate similarity in two KGs from the predicate-proximity-graph, we propose two algorithms for controlling the aggregation of multiple types of an entity and highlighting the most distinctive entity type (e.g., focusing more on \texttt{country} than on \texttt{thing}) via \emph{pseudo-type embedding}.
Such a approximate predicate algorithm provides an automatic way of aligning predicates between two KGs, which not only complements the latent type information but also can be optimized by further joint learning for better predicate embeddings.

To achieve entity alignment, we exploit the attribute triples (in addition to relation triples) for entity alignment and propose \emph{attribute character embeddings} to encode the attribute value to compute attribute embeddings for KGs. Before our work, there was one study that has proposed an embedding for attributes \cite{sun2017cross}. However, it only uses the attribute types for computing embedding, which loses all the content information of the attributes and is ineffective in capturing attribute (dis)similarity. \textit{Hence, we are the first to propose attribute embedding that is based on the textual contents of the attributes \cite{trisedya2019entity}}.
TransAlign generates attribute character embeddings from the attribute triples and then computes the similarity of these attribute embeddings.
The attribute similarity between entities in two KGs helps the attribute embedding to yield a unified embedding space for two KGs. This enables us to use attribute embeddings to shift the entity embeddings of two KGs into the same vector space and hence allows the entity embeddings to capture the similarity between entities from two KGs. 

\textbf{With the above two components, we achieve the first fully automatic method to KG alignment.} The contributions of this paper are as follows.

\begin{enumerate}[label=\textbf{C\arabic*}:] 
	\item We propose TransAlign, a fully automatic entity alignment method that aligns two KGs with no seed alignments required (neither predicate nor entity seed alignments). We make important contributions to two critical components for KG alignment, automatic entity and predicate alignment.
	\item We are the first to propose attribute embedding based on the textual contents of the attributes, which enables automatic entity alignment.
	\item We propose two novel automatic predicate alignment algorithms: (i) we use a predicate-proximity-graph to capture predicates as relationships of entity types, and (ii) we use pseudo-type embeddings that aggregate multiple entity types in the proximity graph as the vector representation for predicates. This way, TransAlign can compute predicate embeddings while simultaneously performing a predicate alignment procedure between two KGs automatically.
	\item We conduct an extensive experimental study, which shows that our method is highly effective while being fully automatic. 
	Compared to existing methods, which all rely on manually crafted seeds, TransAlign outperform the best baseline by up to 1.39\% in hits@10.
\end{enumerate}

This paper is an extended version of our earlier conference paper~\cite{trisedya2019entity}. There, we presented the basic idea of the attribute character embeddings (\textbf{C2}). However, the method in the previous paper~\cite{trisedya2019entity} requires manually crafted predicate alignments. Specifically, it uses edit distance to compute the similarity score between predicates, and manual inspection is required to remove false positives. In this journal extension,
we have made substantial new contributions. First, we propose novel algorithms to align predicates without seed alignments (\textbf{C3}). Second, we provide a scheme for the joint learning of entity, attribute and predicate embeddings, achieving a fully automatic KG alignment method (\textbf{C1}). Third, we have re-run the experiments and conducted a more comprehensive experimental study, including more state-of-the-art
benchmarks such as GNN-based ones (\textbf{C4}).

\section{Related Work} 
	\label{kba-related-work}

We discuss two groups of commonly used entity alignment methods. The first is string-similarity-based methods detailed in Section \ref{related-work-string-similarity-based-entity-alignment}. The second is embedding-based methods detailed in Section \ref{related-work-embedding-based-entity-alignment}.

\subsection{String-Similarity-based Entity Alignment} 
\label{related-work-string-similarity-based-entity-alignment}
Earlier entity alignment methods use string similarity as the main alignment tool. For example, LIMES \cite{ngomo2011limes} first uses the \textit{triangle inequality} to approximate entity similarity. Then, the actual similarity of the entity pairs that have a high approximated similarity is computed, and the entity pair with the highest actual string similarity is returned. RDF-AI \cite{scharffe2009rdf} implements an alignment method that consists of preprocessing, matching, fusion, interlink, and post-processing modules, among which the matching module uses fuzzy string matching based on sequence alignment \cite{rivas1999dynamic}, word relation \cite{fellbaum1998wordnet}, and taxonomical similarity algorithms. SILK \cite{volz2009discovering} allows users to specify the mapping rules using the \textit{Silk - Link Specification Language} (Silk-LSL). SILK provides similarity metrics, including string equality and similarity, numeric similarity, date similarity, and Uniform Resource Identifier (URI) equality.

There are also studies using graph similarity to improve entity alignment. LD-Mapper \cite{raimond2008automatic} combines string similarity with entity nearest neighbor similarity. RuleMiner \cite{niu2012effective} uses an Expectation-Maximization (EM) algorithm to refine a set of manually defined entity matching-rules. HolisticEM \cite{pershina2015holistic} constructs a graph of potential entity pairs based on the overlapping attributes and the neighboring entities. Then, the local and global properties from the graph are propagated using \textit{Personalized Page Rank} (PPR) to compute the actual similarity of entity pairs. The main limitation of the string-similarity-based alignment methods is that they require manually defined entity matching-rules, which determine the attributes to be compared between entities. These rules are error-prone because different pairs of entities may need different attributes to be compared.

\subsection{Embedding-based Entity Alignment} 
\label{related-work-embedding-based-entity-alignment}
Recently, knowledge base embedding methods have been proposed to address KB completion tasks \cite{socher2013reasoning} that aim to predict missing entities (i.e., head entity or tail entity) or relations (i.e., predicates) based on existing triples in a knowledge base. These methods compute a vector representation (i.e., embeddings) of all entities in a KB based on entity nearest neighbors. The embeddings preserve structural information of entities. Hence, entities that share similar neighbors in the KB should have a close vector representation. The embeddings can be used to compute similarities between entities in different KBs for the alignment. In this section, we first review KB embedding methods, then we detail the embedding-based KB alignment methods. 

\subsubsection{Knowledge Base Embedding Methods} 
\label{related-work-kbe-models}

The translation-based method, TransE \cite{bordes2013translating}, is a simple yet effective knowledge base embedding method. TransE represents a relationship between a pair of entities as a translation between the embeddings of the entities. A triple that consists of $\langle \texttt{head, predicate, tail} \rangle$ denoted as $\langle h, p, t \rangle$ is represented as $\mathbf{h + p \approx t}$. This representation indicates that the embedding of the head entity $\mathbf{h}$ is close to the embedding of the tail entity $\mathbf{t}$ by translating $\mathbf{h}$ via the embedding of the predicate $\mathbf{p}$. A scoring function $f(\mathbf{h,t}) = \left\lVert \mathbf{h+p-t}\right\rVert_2$ is used to measure the plausibility of a triple.

In recent years, several studies propose improvements over TransE. TransH \cite{wang2014knowledge} uses a normalization vector to project the embeddings of the head entity and tail entity into a hyperplane for handling N-to-1, 1-to-N, and N-to-N relationships. TransR \cite{lin2015learning} further improves TransH by separating the relationship (i.e., predicate) vector space from the entity vector space to improve the coherence between the learned embeddings. TransSparse \cite{ji2016knowledge} handles unbalanced relationships in a knowledge graph. Some relationships (i.e., predicates) connect many entity pairs while others only connect a few. To address this problem, TransSparse uses a controlled projection matrix with a sparse degree $\theta$ determined by the number of entity pairs connected by the corresponding relationships used to normalize the imbalance relationship.

Another popular paradigm of KG embedding is via graph neural networks, such as Graph Convolutional Networks (GCN)~\cite{kipf2017semi} and Graph Attention Networks (GAT)~\cite{velickovic2018graph}. Recently, graph embedding based on Transfomers~\cite{yun2019graph} have been proposed such as  HGT~\cite{hu2020heterogeneous} and RHGT~\cite{mei2022relation}. These methods learn entity embeddings via information propagation between nodes in a graph. 

\subsubsection{Knowledge Base Alignment Methods} 
\label{related-work-kba-models}

The embedding methods above aim to preserve the structural information of the entities, i.e., entities that share similar neighbor structures in the KB should have a close representation in the embedding space. The advancement of such embedding methods motivates researchers to study embedding-based entity alignment. Chen et al. \cite{chen2017multilingual} propose MTransE, an embedding-based method for multilingual entity alignment based on TransE. MTransE uses a \textit{knowledge module} and an \textit{alignment module} to learn the multilingual knowledge graph structure. The knowledge module is a standard entity embedding method (i.e., TransE) that computes entity and relationship embeddings of a knowledge base. The alignment module learns a matrix to translate both entity and relationships from different embedding spaces (i.e., different KBs) into a unified embedding space. For computing the transition matrix, they propose three algorithms: (1) \textit{distance-based axis calibration} unit that  penalizes the alignment based on the distances of cross-lingual counterpart; (2) \textit{translation vectors} unit that encodes cross-lingual transitions into a vector and considers it as an additional translation to compute the plausibility score; and (3) \textit{linear transformations} unit that uses additional neural network layers to transform the entity and relationship embeddings. In the follow-up work, Chen et al. \cite{chen2017multigraph} propose a generalized affine-map-based method to improve the alignment method of MTransE for handling various forms of invertible transformations, such as translation and scaling.

Zhu et al. \cite{zhu2017iterative} propose an iterative method for entity alignment via joint embeddings. Their method consists of three modules. The first is a \textit{knowledge embedding} module that learns entity and relationship embeddings for each knowledge base. The second is a \textit{joint embedding} module that maps the entity and relationship embeddings from different KBs into a joint semantic space according to a seed set of known aligned entities. The third is an \textit{iterative alignment} module that updates the entity and relationship embeddings by taking the high-confidence aligned entities found in the previous iteration as additional seed alignments.

Sun et al. \cite{sun2017cross} propose a joint attribute embedding method (called JAPE) for cross-lingual entity alignment. JAPE improves the alignment by capturing the correlations of attributes via the attribute type similarity. This method differs from ours in terms of how the attributes are exploited. In JAPE, the attribute values are replaced by their corresponding primitive types (e.g., \emph{integer}, \emph{string}, etc.). In contrast, our method uses the actual attribute values for computing attribute embeddings.

As the other popular paradigm for KG embedding, many GNN-based entity alignment methods have been proposed \cite{cao2019multi,sun2020knowledge,xu2019cross,wang2018cross,wu2019relation,ye2019a}. 
GNN-based methods, such as Graph Attention Network (GAT) \cite{velickovic2018graph} and Graph Convolutional Network (GCN)\cite{kipf2017semi}, compute each entity's embedding by aggregating the entity's neighbors' embeddings according to the graph structure; the manually crafted seed alignments impose that the pair of aligned entities in the two KGs must have the same embedding during the learning process, which make the vector spaces of the two KGs aligned.
To improve the performance of GNN for entity alignment, existing methods \cite{cao2019multi,wang2018cross} combine entity embeddings and attribute type embeddings in the computation of GNN. 
More recently, entity alignment based on graph-transfomers have been proposed \cite{cai2022entity}. 
However, all the above methods still require manual work.

In summary, all existing embedding-based entity alignment methods require a set of manually crafted seed alignments. There has been no existing work on automatic predicate alignment. To address these problems, this paper introduces a novel joint learning scheme of entity, predicate, and attribute embeddings for entity alignment, which does not require any manually crafted seed alignments.

\section{Preliminary} 
	\label{kba-preliminary}
	
We start with the problem definition. A knowledge graph $\mathcal{G}$ consists of a combination of relation triples and attribute triples. A relationship triple is in the form of $\left<h,p,t\right>$, where $p$ is a relationship (\textit{predicate}) between two entities $h$~(\textit{head}) and $t$ (\textit{tail}). An attribute triple is in the form of $\left<h,p,v\right>$ where $v$ is an attribute value of entity $h$ with respect to the predicate $p$. 

Given two knowledge graphs $\mathcal{G}_1$ and $\mathcal{G}_2$, the task of entity alignment aims to find every pair $\langle h_1, h_2 \rangle$ where $h_1 \in \mathcal{G}_1$, $h_2 \in \mathcal{G}_2$, and $h_1$ and $h_2$ represent the same real-world entity. We use an embedding-based method that assigns a continuous representation for each element of a triple in the forms of $\mathbf{\left<h,p,t\right>}$ and $\mathbf{\left<h,p,v\right>}$, where the bold-face letters denote the vector representations of the corresponding element.

Our proposed method is built on top of a translation-based embedding method. We first discuss translation-based embedding methods and their limitations when being used for entity alignment before presenting our method.

\begin{figure*}[t!]
	\begin{center}
		\includegraphics[width=.9\textwidth]{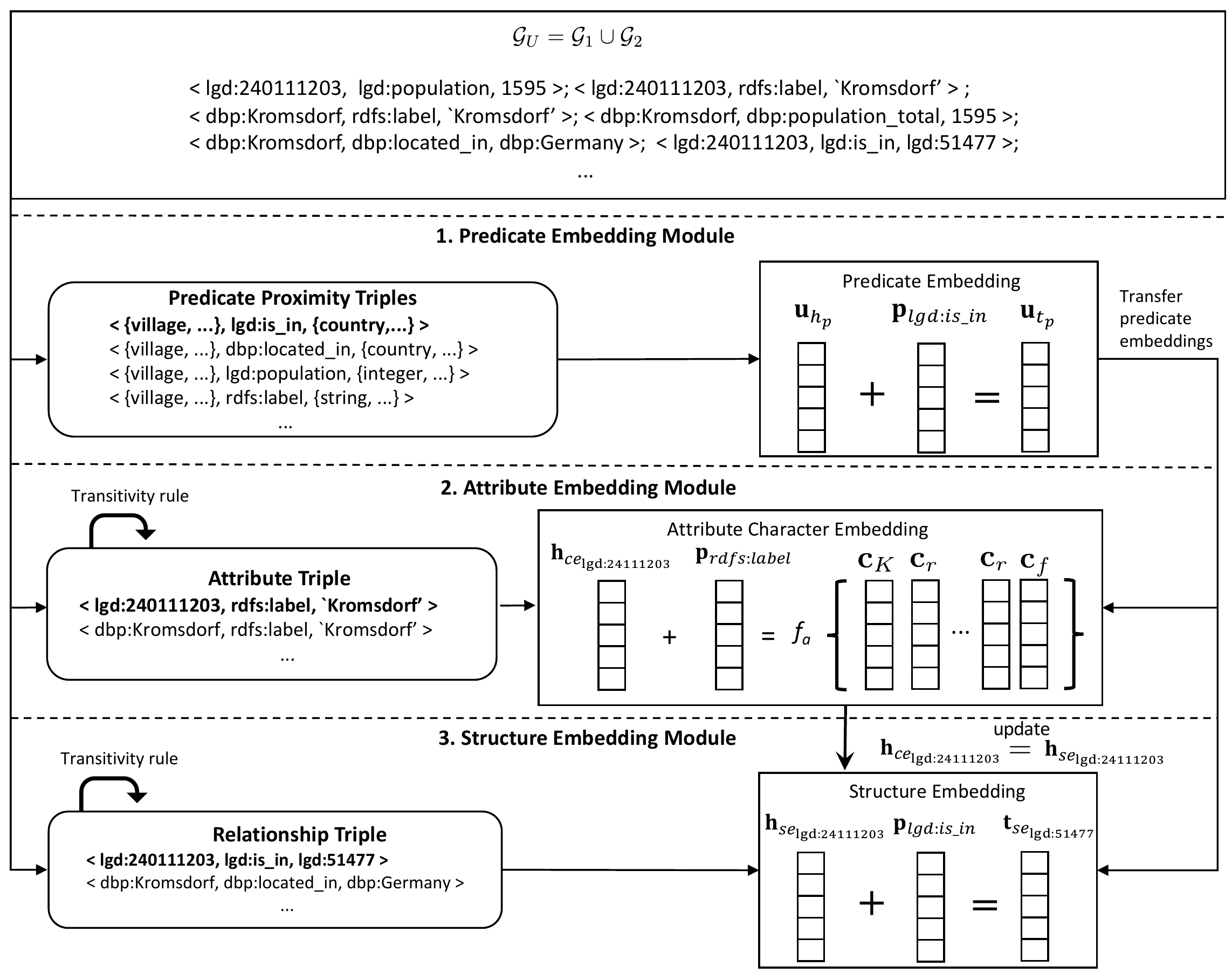}
	\end{center}
	\vspace{-3mm}
	\caption{\label{fig-kba-overall-module} Overview of our proposed TransAlign method for entity alignment.}
\end{figure*}

\subsection{Translation-based Embedding Method} 
	\label{kba-transe}

Given a relationship triple $\langle h,p,t \rangle$, a translation-based embedding method, such as TransE~\cite{bordes2013translating}, suggests that the embedding of the tail entity $t$ should be close to the embedding of the head entity $h$ plus the embedding of the relationship $p$, i.e., $\mathbf{h + p \approx t}$. Such an embedding method aims to preserve the structural information of the entities, i.e., entities that share similar neighbor structures in a knowledge graph should have a close representation in the embedding space. We call it the \emph{structure embedding}. To learn the structure embedding, TransE minimizes a margin-based objective function $\mathcal{J}_{SE}$:
\begin{align}
	\mathcal{J}_{SE}&=\sum_{t_r\in \mathcal{T}_r} \sum_{t_r^\prime \in \mathcal{T}_r^\prime} \max\left(0,\left[ \gamma+  f(t_r)- f(t_r^\prime) \right] \right) \label{eq-kba-ori-jse}\\
	f(t_r)&=\left\Vert \mathbf{h+p-t} \right\Vert_2 \\
	\mathcal{T}_r&=\{\langle h, p, t \rangle | \langle h,p,t \rangle \in \mathcal{G}\} \\
	{\mathcal{T}_r}^\prime&=\left\{\left<h^\prime,p,t\right>\left.\right|h^\prime\in \mathcal{E}\right\} \cup\left\{\left< h,p,t^\prime\right>\left.\right|t^\prime\in \mathcal{E}\right\}
\end{align}
Here, $\left\Vert \textbf{x} \right\Vert_2$ is the L2-Norm of vector $\textbf{x}$,  $\gamma$ is a margin hyperparameter, $\mathcal{T}_r$ is the set of valid relation triples, and $\mathcal{T}_r^\prime$ is the set of corrupted relation triples ($\mathcal{E}$ is the set of entities in $\mathcal{G}$). The corrupted triples are used as negative samples created by replacing the head or tail entity of a valid triple in $\mathcal{T}_r$ with a random entity.

\begin{figure*}[t!]
	\centering
	\includegraphics[width=.9\textwidth]{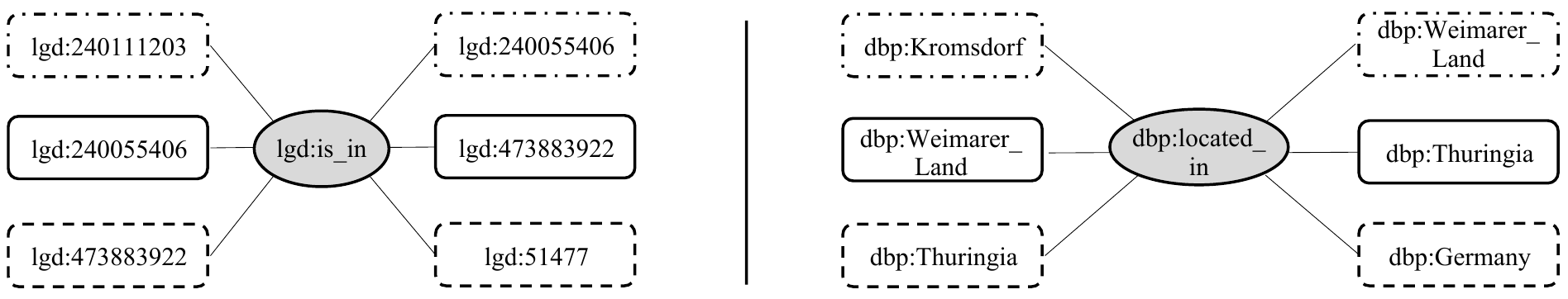}
	\caption{Predicate graph of the similar relationship \texttt{lgd:is\_in} (left) and \texttt{dbp:located\_in} (right) in two KGs. Each type of dotted line represents the similar entity types in two predicate graphs. }
	\label{fig-kba-predicate-graph-raw}
	\vspace{-3mm}
\end{figure*}

The advantages of structure embeddings drive further studies of embedding-based entity alignment. However, a straightforward implementation of structure embedding for entity alignment has limitations: the entity embeddings computed on different KGs may fall in different spaces, where similarity cannot be computed directly. Existing methods~\cite{chen2017multilingual,sun2017cross,zhu2017iterative} address this limitation by computing a transition matrix to map the embedding spaces of different KGs into the same space, as discussed earlier. However, such methods require manually collecting a seed set of aligned entities from the different KGs to compute the transition matrix, which does not scale and is vulnerable to the quality of the selected seed aligned entities.

Next, we detail our method to address these limitations.

\section{Proposed Method} 
	\label{kba-proposed-model}
We present an overview of \emph{TransAlign} in Section \ref{kba-solution-overview}. We detail the components of TransAlign afterwards, including \emph{predicate embedding} module in Section \ref{kba-predicate-embedding}, \emph{structure embedding} module in Section \ref{kba-structure-embedding}, \emph{attribute embedding} module in Section \ref{kba-attribute-character-embedding}, joint learning 
scheme in Section \ref{kba-joint-learning}, entity alignment in Section \ref{kba-entity-alignment}, and triple enrichment in Section~\ref{kba-transitivity-rule}.

\subsection{TransAlign Overview} 
	\label{kba-solution-overview}
TransAlign is a fully automatic and effective embedding-based entity alignment method for aligning entities across different KGs. It jointly learns entity, predicate, and attribute embeddings from different KGs. TransAlign consists of three embedding modules, including \textit{predicate embedding}, \textit{attribute embedding}, and \textit{structure embedding}. Fig.~\ref{fig-kba-overall-module} illustrates the interaction of these embedding modules in TransAlign.

Embedding-based entity alignment requires the embeddings (both predicate and entity embeddings) of two KGs to fall in the same vector space. To obtain such embeddings, we first combine two knowledge graphs. We denote the combined graph as $\mathcal{G}_U = \mathcal{G}_1 \cup \mathcal{G}_2$. From the combined graph, we create three sets of triples, including the set of predicate-proximity triples $\mathcal{T}_p$, the set of relation triples $\mathcal{T}_r$, and the set of attribute triples $\mathcal{T}_a$. In the set of predicate-proximity triples, each predicate represents a relationship between entity types. For example, the set of predicate-proximity triples may contain the triples $\langle$\texttt{village}, \texttt{dbp:located\_in}, \texttt{country}$\rangle$ and $\langle$\texttt{village}, \texttt{lgd:is\_in}, \texttt{country}$\rangle$, which are used to train a predicate embedding module (detailed in Section \ref{kba-predicate-embedding}) for capturing the similarity between predicates from two KGs, e.g., \texttt{dbp:located\_in} and \texttt{lgd:is\_in}. This way, TransAlign not only generates a unified embedding space of predicates but also captures the similarity between predicates from two KGs. The predicate embeddings are then used to compute the attribute and the structure embeddings.

The structure (i.e., entity) embedding (detailed in Section \ref{kba-structure-embedding}) is learned using the set of relation triples $\mathcal{T}_r$, while the attribute embedding (detailed in Section \ref{kba-attribute-character-embedding}) is learned using the set of attribute triples $\mathcal{T}_a$. Initially, the entity embeddings of the entities from $\mathcal{G}_1$ and $\mathcal{G}_2$ fall into different vector spaces because the entities from both KGs are represented using different naming schemes. On the contrary, the attribute embeddings learned from the attribute triples $\mathcal{T}_a$ can fall into the same vector space. This is achieved by learning character embeddings from the attribute strings, which can be similar even if the attributes are from different KGs (we call it \textit{attribute embedding}). Then, we use the learned attribute embeddings to shift the entity embeddings of the entities into the same vector space, which enables the entity embeddings to capture the similarity between entities from two KGs. As an example, suppose that we have triples $\langle \texttt{lgd:240111203, lgd:is\_in, lgd:51477} \rangle$ and $\langle$\texttt{lgd:51477, rdfs:label, "Germany"}$\rangle$  from $\mathcal{G}_1$, and $\langle$\texttt{dbp:Kromsdorf, dbp:located\_in, dbp:Germany}$\rangle$ and  $\langle\texttt{dbp:Germany, rdfs:label, "Germany"}\rangle$ from $\mathcal{G}_2$. The attribute embedding allows both entities \texttt{lgd:5147} and \texttt{dbp:Germany} to have similar vector representations since both entities have a similar attribute value \texttt{"Germany"}. Then, the structure embeddings of entities \texttt{lgd:240111203} and \texttt{dbp:Kromsdorf} will also be similar since they share similar predicate representation (from the predicate embedding similarity) and have two tail entities \texttt{lgd:51477} and \texttt{dbp:Germany}, which have similar vector representations. 

Once we have the embeddings for all entities in $\mathcal{G}_1$ and $\mathcal{G}_2$, the entity alignment module (detailed in Section \ref{kba-entity-alignment}) finds every pair $\left<h_1, h_2\right>$ where $h_1 \in \mathcal{G}_1$ and $h_2 \in \mathcal{G}_2$ with a similarity score above a threshold $\beta$. 

To further improve the performance of TransAlign, we use the transitivity rule to enrich the properties of an entity that helps build a more robust attribute embedding for computing the similarity between entities. This is detailed in Section \ref{kba-transitivity-rule}.

\subsection{Predicate Embedding Module} 
	\label{kba-predicate-embedding}
	
The same predicates from two knowledge graphs typically connect the same real-world entities. However, they may exist in different surface forms. For the example in Fig.~\ref{fig-kba-predicate-graph-raw}, the predicate \texttt{lgd:is\_in} in LinkedGeoData and the predicate \texttt{dbp:located\_in} in DBpedia connect three entity pairs. In LinkedGeoData, the predicate \texttt{lgd:is\_in} connects $\langle$\texttt{lgd:240111203,lgd:240055406}$\rangle$, $\langle$\texttt{lgd:240055406, lgd:473883922}$\rangle$, and $\langle$\texttt{lgd:473883922,lgd:51477}$\rangle$. Meanwhile, the predicate \texttt{dbp:located\_in} in DBpedia connects $\langle$\texttt{dbp:Kromsdorf,dbp:Weimarer\_Land}$\rangle$, $\langle$\texttt{dbp:Weimarer\_Land,dbp:Thuringia}$\rangle$, and $\langle$\texttt{dbp:-\\Thuringia,dbp:Germany}$\rangle$. Here, the entity pairs from both knowledge graphs correspond to the same real-world entity pairs. For example, the head and tail entities of the entity pair $\langle$\texttt{lgd:240111203,lgd:240055406}$\rangle$ and $\langle$\texttt{dbp:Kromsdorf,dbp:Weimarer\_Land}$\rangle$ correspond to a village named Kromsdorf and a district named Weimarer Land, respectively. However, due to the different naming schemes of two knowledge graphs, entity embedding methods may not capture this similarity. If we apply an entity embedding 
method to the raw knowledge graph, the predicate embeddings of \texttt{lgd:is\_in} and \texttt{dbp:located\_in} may fall in different vector spaces.

In our previously proposed method \cite{trisedya2019entity}, we use a semi-automatic predicate alignment procedure to handle the aforementioned problem. We rename the similar predicates of two KGs with a unified naming scheme to have a unified vector space for the relationship embeddings. First, we use string edit distance to find similar predicates from two KGs, and then we manually remove the false positives. However, this procedure is unpractical in real applications as it requires human intervention to align predicates (manually crafted seed alignments) from two KGs. 

To address the above problem and provide a fully automatic predicate alignment procedure, TransAlign learns predicate embeddings from a \emph{predicate-proximity-graph} of two KGs. Our method can automatically capture the similarity between predicates across two KGs. A predicate-proximity-graph is a graph that represents the relationships between \emph{entity types} instead of \emph{entities}. We create the predicate-proximity-graph by replacing entities by their corresponding types. First, we merge two knowledge graphs to form a combined graph $\mathcal{G}_U = \mathcal{G}_1 \cup \mathcal{G}_2$. Then, we replace the head entity and tail entity of each triple in the combined graph to their corresponding entity types.

We obtain the entity types by extracting the value of \texttt{rdfs:type} predicate for all entities from each KG. Typically, each entity has multiple types. For example, the entity \texttt{Germany} may have multiple entity types \{\texttt{thing, place, location, country}\} in a KG. Moreover, different KGs may have different schemes of entity types, e.g., in another KG, the entity \texttt{Germany} may have entity types \{\texttt{place, country}\}. To address these issues, in the predicate-proximity-graph, we replace the head entity and the tail entity of each triple in a knowledge graph with a set of entity types. For example, we replace the triples $\langle$\texttt{dbp:Kromsdorf}, \texttt{dbp:located\_in}, \texttt{dbp:Germany}$\rangle$ with the triples $\langle$$\mathcal{U}_{kromsdorf}$, \texttt{dbp:located\_in}, $\mathcal{U}_{germany}$$\rangle$. Here, $\mathcal{U}_{x}$ is a set of types for entity $x$, e.g., $\mathcal{U}_{germany}$= \{\texttt{thing, place, location, country}\}.

To capture the predicate similarity in two KGs, the predicate embedding module should focus on the most distinctive entity type, e.g., focusing on \texttt{country} more than on \texttt{thing}. We propose two algorithms aggregating multiple entity types: i) weighted sum function, 2) attention-based function.


\textbf{Weighted Sum Function:} Given entity type embedding $\mathbf{U}_x=(\vec{\mathbf{z}}_0, \vec{\mathbf{z}}_1, \cdots, \vec{\mathbf{z}}_M)$ of entity type $z \in \mathcal{U}_x $ with $M$ types, we calculate the \emph{pseudo-type embedding} (denoted as $\mathbf{u}$) as follows.
%
\begin{equation}
    \label{eq-weight-u}
    \mathbf{u}=\sum_{i=0}^{M} w_i \vec{\mathbf{z}}_i
\end{equation}
\begin{equation}
    w_i = \text{softmax}\left(\frac{l_i}{a_i r_i}{k_i}\right)
\end{equation}
Here, the weight $w_i$ controls the distributions of $\vec{\mathbf{z}}_i$, which is the vector representation of an entity type $z$ in $\mathcal{U}_x$. To give a larger weight for the most distinctive type, we use $l_i$, which is the level of specificity of an entity type in WordNet \cite{fellbaum1998wordnet}. An entity type with a deeper level of specificity has a larger value of $l_i$, e.g., \texttt{country} has a deeper level of specificity than \texttt{thing}. This way, the predicate embedding module can highlight the most distinct type. We normalize the weight using three variables. The first is $a_i$, which is the number of attributes in $\mathcal{U}_x$. The second is $r_i$, which is the number of occurrences of type $z$ in a KG. Intuitively, an entity type that appears in almost all entities such as \texttt{thing} is less distinctive, and the predicate embedding module should filter this entity type to obtain a better predicate representation. The third variable is $k_i$, which is the number of KGs that contains the entity type $z$ to indicate the agreements between two KGs in the entity type $z$. Lastly, we use $\text{softmax}$ to transform the weights into the probability distributions of each type in $\mathcal{U}_x$.

\textbf{Attention-based Function:} Contrary to the most distinctive entity types, there may be some ``noise'' entity types which contributes little when learning the meaning of a predicate. For example, in the set of entity types for entity \texttt{Germany}, $\mathcal{U}_{germany}$= \{\texttt{thing, place, location, country}\}, the entity type \texttt{thing} $\in \mathcal{U}_{germany}$ may be less relevant to the triple $\langle$\texttt{dbp:Kromsdorf}, \texttt{dbp:located\_in}, \texttt{dbp:Germany}$\rangle$, i.e.,  \texttt{thing} can be seen as a ``noise'' entity type which is not essential for learning the representation of predicate \texttt{dbp:located\_in}. 

To further capture the importance of different entity types for a predicate, we propose an attention-based algorithm, which helps the module
adaptively evaluate the entity type weight and ignore the potential type noise.
We calculate the attention weight $z_i$ of the $i$-th entity type as:  
\begin{equation}
z_i=softmax(U^T_x W_z  \vec{\mathbf{z}}_i)
\end{equation}
where $W_z$ denotes the trainable weight matrix of entity type embedding $ \vec{\mathbf{z}}_i$. The final pseudo-type embedding $\mathbf{u}$ is obtained through a weighted sum of all corresponding entity type vectors $ \vec{\mathbf{z}}_i$ representing different semantic meaning:
\begin{equation}
\label{eq-attention-u}
\mathbf{u}=\sum_{i=0}^{M}z_i  \vec{\mathbf{z}}_i
\end{equation}
where $ \vec{\mathbf{z}}_i$ is the embedding of the $i$-th entity type. 

The pseudo-type embeddings computed by Equations \ref{eq-weight-u} or \ref{eq-attention-u} are used as the proximity entity embeddings, which are used next to train the predicate embeddings as follows. 

We use the pseudo-type embeddings $\mathbf{u_{h_{p}}}$ and $\mathbf{u_{t_{p}}}$ to represent the corresponding head entity and tail entity in the predicate proximity triples $\mathcal{T}_p$ respectively. We then compute the predicate embeddings by minimizing the following objective function:
\begin{eqnarray}
\mathcal{J}_{PE}=\sum_{t_p\in \mathcal{T}_p} \sum_{t_p^\prime \in \mathcal{T}_p^\prime} \max\left(0,\left[ \gamma+  f(t_p)- f(t_p^\prime) \right] \right)\\
f(t_p)=\left\Vert \mathbf{u_{h_{p}}+p-u_{t_{p}}} \right\Vert_2
\label{eq-J-PE}
\end{eqnarray}
where $t_p$ is a triple in the predicate-proximity-graph and $t_p^\prime$ is a corrupted triple (i.e., for negative samples) generated based on the predicate-proximity-graph. Here $\mathbf{u_{h_{p}}}$ and $\mathbf{u_{t_{p}}}$ can be obtained by the above two functions, Eq. \ref{eq-weight-u} and Eq. \ref{eq-attention-u}. 
We use TransAlign-W and TransAlign-A to denote TransAlign using Eq. \ref{eq-weight-u} and Eq. \ref{eq-attention-u}, respectively.

Note that the procedure above can also be used to compute embeddings for attribute predicates. This is achieved by adapting Eq. \ref{eq-J-PE} to replace the entity types of the tail entity in relation triples with literal types (e.g., string, integer and long data type) of attribute values in attribute triples.

By optimizing the above objective function, TransAlign yields a unified predicate embedding space from two knowledge graphs. We compute the predicate embeddings from two KGs, and hence we can transfer these unified predicate embeddings to learn the structure and the attribute embeddings.

\subsection{Structure Embedding Module} 
\label{kba-structure-embedding}

We follow TransE to compute the structure embeddings. TransE uses the same weight for each neighbor in computing the embeddings of an entity, i.e., to update the embeddings of an entity, all neighbors have the same contribution. We adapt TransE to give different weights to different neighbors of an entity. Our intuition is to give larger weights to neighbors connected by predicates that are already aligned, which provide a good signal for entity alignment.


From the example in Table \ref{table-kba-example}, we can group the predicates of KGs into three groups. The first group is the already aligned predicates, such as \texttt{geo:lat, geo:long,} and \texttt{rdfs:label}. These predicates follow the predicate naming scheme convention\footnote{https://www.w3.org/TR/rdf-schema/} in knowledge bases. The second group is the implicitly aligned predicates, such as \texttt{lgd:is\_in} and \texttt{dbp:located\_in}. These predicates are helpful for entity alignment if we can capture the alignment between these predicates. We handle this problem by our predicate embedding module (cf. Section~\ref{kba-predicate-embedding}). The last group is the non-aligned predicates, such as the predicates \texttt{lgd:alderman} and \texttt{dbp:district}. These predicates do not help in entity alignment and are considered as noises.

To reduce the effect of the noises, we adapt TransE by adding a weight $\alpha$ to control the embedding learning over the triples. Hence, the entity embedding method can filter the non-aligned triples based on the non-aligned predicates. To learn the structure embedding, in TransAlign, we minimize the objective function $\mathcal{J}_{SE}$ adapted from Eq.~\eqref{eq-kba-ori-jse} as follows:
\begin{align}
\mathcal{J}_{SE}&=\sum_{t_r\in\ \mathcal{T}_r} \sum_{t_r^\prime \in\ \mathcal{T}_r^\prime} \max\left(0,\gamma+ \alpha \left( f(t_r)- f(t_r^\prime) \right) \right)\label{eq-kba-jse}\\
\alpha&=\frac{count(r)}{\left| \mathcal{T} \right|}
\end{align}
where $\mathcal{T}_r$ is the set of valid relation triples, $\mathcal{T}_r^\prime$ is the set of corrupted relation triples, $count(r)$ is the number of occurrences of relationship $r$, and $\left| \mathcal{T} \right|$ is the total number of triples in the merge KG $\mathcal{G}_{U}$. Typically, the number of occurrences of the already aligned and the implicitly aligned predicates is larger than that of the non-aligned predicates (since the aligned predicates appear in both KGs while the non-aligned predicates only appear in one of the KGs). Hence, our weighting algorithm allows the embedding method to learn more from the aligned triples. For example, for the triples in Table \ref{table-kba-example}, the weight $\alpha$ helps the embedding method to focus more on relationships \texttt{rdfs:label}, \texttt{geo:lat}, and \texttt{geo:long} ($\alpha = 2/12$ for each of these predicates) than on relationships \texttt{lgd:alderman} or \texttt{dbp:district} ($\alpha = 1/12$ for each of these predicates).

\subsection{Attribute Embedding Module} 
\label{kba-attribute-character-embedding}

For the attribute embedding, we interpret attribute predicate $p$ as a translation from the head entity $h$ to the attribute value $v$. The same attribute may appear in different forms in two KGs, e.g., \texttt{50.9989} vs. \texttt{50.9988888889} as the latitude of an entity; \texttt{"Barack Obama"} vs. \texttt{"Barack Hussein Obama"} as a person name, etc. Hence, we use a compositional function to encode the attribute value and define the relationship of each element in an attribute triple as $\mathbf{h}~+~\mathbf{p}~\approx~f_a(v)$. Here, $f_a(v)$ is a compositional function and $v$ is a sequence of the characters of the attribute value $v~=~\left\{ c_1,c_2,c_3,..., c_t \right\}$. The compositional function encodes the attribute value into a single vector and maps similar attribute values to a similar vector representation. We consider three compositional functions as follows.

\textbf{Sum Compositional Function} (\textbf{SUM}). The first compositional function is defined as a summation of all character embeddings of the attribute value.
\begin{equation}
f_a(v) = \mathbf{c_1 + c_2 + c_3 + ... + c_t}
\end{equation}
where $\mathbf{c_1, c_2, ..., c_t}$ are the character embeddings of the attribute value. This compositional function is simple, but it suffers in that two strings that contain the same set of characters with a different order will have the same vector representation (i.e., order invariant). For example, two coordinates, \texttt{"50.15"} and \texttt{"15.05"}, will have the same vector representation.

\textbf{LSTM-based Compositional Function} (\textbf{LSTM}). To address the problem of SUM, we propose an LSTM-based compositional function. This function uses LSTM networks to encode a sequence of characters into a single vector. We use the final hidden state of the LSTM networks as a vector representation of the attribute value.
\begin{equation}
f_a(v) = f_{lstm}({\mathbf{c_1, c_2, c_3, ..., c_t}})
\end{equation}
where $f_{lstm}$ is an LSTM network \cite{hochreiter1997long}.

\textbf{N-gram-based Compositional Function} (\textbf{N-gram}). LSTM-based compositional function handles the order invariant problem. However, it only considers the unigram features of a string. To capture rich compositional information of a string, we further propose an N-gram-based compositional function as an alternative to the above two compositional functions. Here, we use the summation of the n-gram combination of the attribute value.
\begin{equation}
f_a(v)=\sum_{n=1}^{N}\left(\frac{\sum_{i=1}^{l}\sum_{j=i}^{n}\mathbf{c_j}}{t-i-1}\right)
\end{equation}
where $N$ indicates the maximum value of $n$ used in the n-gram combinations ($N = 10$ in our experiments), and $l$ is the length of the attribute value.

To learn the attribute embedding, we minimize the following objective function $\mathcal{J}_{CE}$:
\begin{align}
\mathcal{J}_{CE}&=\sum_{t_a\in \mathcal{T}_a} \sum_{t_a^\prime \in \mathcal{T}_a^\prime} \max\left(0,\left[ \gamma+  \alpha \left( f(t_a)- f(t_a^\prime) \right) \right] \right)\label{eq-kba-jce}\\
f(t_a)&=\left\Vert \mathbf{h}+\mathbf{p}-f_a(v) \right\Vert_2\\
\mathcal{T}_a&=\{\langle h,p,v \rangle \in \mathcal{G}_U\}\\
{\mathcal{T}_a}^\prime&=\left\{\left.\left<h^\prime,p,v\right>\right|h^\prime\in \mathcal{E}_U\right\}\cup\left\{\left<h,p,v^\prime\right>|v^\prime\in \mathcal{A}_U\right\}
\end{align}
Here, $\mathcal{T}_a$ is the set of valid attribute triples from the training dataset, and $\mathcal{T}_a^\prime$ is the set of corrupted attribute triples ($\mathcal{A}_U$ is the set of attributes in $\mathcal{G}_U$). The corrupted triples are used as negative samples by replacing the head entity with a random entity or the attribute with a random attribute value. $f(t_a)$ is the plausibility score computed based on the embedding of the head entity $h$, the embedding of the attribute predicate $p$, and the vector representation of the attribute value computed using function $f_a(v)$.

\subsection{Joint Learning Embedding Scheme} 
\label{kba-joint-learning}

TransAlign jointly learns the predicate embeddings, the structure embeddings, and the attribute embeddings. The proposed method first trains over the predicate-proximity-graph to yield the unified predicate embedding space. 
TransAlign then uses these predicate embeddings to jointly learn the structure and attribute embeddings.
However, the attribute embedding module yields a unified embedding space for two knowledge graphs but lacks structure information. On the other hand, the structure embedding module may yield different embedding space for two knowledge graphs. Thus, we use the attribute embedding $\mathbf{h_{ce}}$ to shift the structure embedding $\mathbf{h_{se}}$ into the same vector space by minimizing the following objective function $\mathcal{J}_{SIM}$:
\begin{equation}
\mathcal{J}_{SIM}=\sum_{s \in \mathcal{G}_1 \cup \mathcal{G}_2} \left[1- \cos(\mathbf{h_{se}},\mathbf{h_{ce}}) \right]
\end{equation}
Here, $\cos(\mathbf{h_{se}},\mathbf{h_{ce}})$ is the cosine similarity of vector $\mathbf{h_{se}}$ and $\mathbf{h_{ce}}$. As a result, the structure embedding captures the similarity of entities between two KGs based on entity relationships, while the attribute embedding captures the similarity of entities based on attribute values. The overall objective function of the joint learning is:
\begin{equation}
\mathcal{J} = \mathcal{J}_{PE} + \mathcal{J}_{SE} + \mathcal{J}_{CE} + \mathcal{J}_{SIM}
\end{equation}

\subsection{Entity Alignment} 
\label{kba-entity-alignment}

The existing embedding-based entity alignment methods are supervised when obtain the resulting embeddings since they need seed alignments to learn entity alignments from two knowledge graphs. Unlike the existing methods, TransAlign captures the similarity between entities from two knowledge graphs by learning a unified entity embedding space via predicate and attribute embeddings. TransAlign does not need seed alignments. Our joint learning embedding scheme lets similar entities from $\mathcal{G}_1$ and $\mathcal{G}_2$ have close vector representations. Thus, the resultant embeddings can be used for entity alignment. We compute the following equation for entity alignment.
\begin{equation}
h_{map}=\argmax_{h_2 \in \mathcal{G}_2} \cos(\mathbf{h}_1, \mathbf{h}_2) \label{eq-kba-hmap}
\end{equation}
Given an entity $h_1 \in \mathcal{G}_1$, we compute the similarity between $h_1$ and all entities $h_2 \in \mathcal{G}_2$; $\left< h_1, h_{map} \right>$ is the expected pair of aligned entities. We use a similarity threshold $\beta$ to filter the pairs of entities that are too dissimilar to be aligned.

\subsection{Triple Enrichment via Transitivity Rule}
\label{kba-transitivity-rule}

In translation-based embedding methods such as TransE, the embedding of an entity is learned by aggregating information from its immediate neighbors (i.e., one-hop neighbors). These methods may implicitly learn the multi-hop relationships between entities via information propagation after many training epochs. However, the information propagation of the multi-hop relationship is weak. On the other hand, the explicit inclusion of multi-hop relationships (e.g., transitive relationships) increases the number of attributes and related entities for each entity, which helps identify the similarity between entities. For example, given triples $\langle$\texttt{dbp:Emporium\_Tower, :locatedIn, dbp:London}$\rangle$ and $\langle$\texttt{dbp:London, :country, dbp:England}$\rangle$, we can infer that \texttt{dbp:Emporium\_Tower} has a relationship (i.e., \texttt{":locatedInCountry"}) with \texttt{dbp:England}. In fact, this information can be used to enrich the related entity \texttt{dbp:Emporium\_Tower}. We treat the one-hop transitive relation as follows. Given transitive triples $\left< h_1,p_1, t_1 \right>$ and $\left< t_1, p_2, t_2 \right>$, we interpret $p_1.p_2$ as a relation from head entity $h_1$ to tail entity $t_2$. Therefore, the relationship between these transitive triples is defined as $\mathbf{h_1 + (p_1 . p_2) \approx t_2}$. The objective functions of the transitivity-enhanced embedding methods are adapted from the Eq. \eqref{eq-kba-jse} and Eq. \eqref{eq-kba-jce} by replacing the relationship vector $\mathbf{p}$ with $\mathbf{p_1.p_2}$.

\section{Experiments} 
\label{kba-experiments}

We perform experiments to show the effectiveness of TransAlign from three different aspects. First, we show the accuracy of TransAlign in entity alignment, which is the main task in this paper. Second, we show that our predicate embedding module achieves better performance on aligning predicates from different knowledge graphs. Third, we show that the resulting embeddings of TransAlign preserve the structure information of knowledge graphs, enabling them to be used in broader applications such as KG completion.
\vspace{-3mm}

\subsection{Datasets}
\label{kba-experiments-dataset}
\begin{table}[t!]
	\centering
	\caption{Statistics of the datasets for entity alignment.}
	\label{table-dataset_statistics}
	\resizebox{0.97\columnwidth}{!}{%
		\begin{tabular}{l|c|c|c|c}
			\toprule[2pt]
			\midrule
			\multicolumn{1}{c|}{Subset} & \multicolumn{1}{p{3.57em}|}{Unique\newline{}entities} & \multicolumn{1}{l|}{Predicates} & \multicolumn{1}{p{5.715em}|}{Relationship\newline{}triples} & \multicolumn{1}{p{4.215em}}{Attribute\newline{}triples} \\
			\midrule
			\midrule
			\multicolumn{5}{c}{\textbf{DW-NB}} \\
			\midrule
			DBpedia & 84,911 &  545   & 203,502 & 221,591 \\
			Wikidata & 86,116 & 703   & 198,797 & 223,232 \\
			\midrule
			\midrule
			\multicolumn{5}{c}{\textbf{DY-NB}} \\
			\midrule
			DBpedia & 58,858 &  211   & 87,676 & 173,520 \\
			Yago  & 60,228 & 91    & 66,546 & 186,328 \\
			\midrule
			\bottomrule[2pt]
		\end{tabular}%
	}
	\vspace*{-3mm}
\end{table}%
We evaluate our method on the latest comprehensive benchmark for KG alignment, \emph{DWY-NB} \cite{zhang2022benchmark}, which consists of two datasets \emph{DW-NB} and \emph{DY-NB}.
The two KGs of DW-NB are subsets of DBpedia~\cite{auer2007dbpedia} and Wikidata~\cite{vrandecick2014wikidata}, respectively. The two KGs of DY-NB are subsets of DBpedia~\cite{auer2007dbpedia} and Yago~\cite{hoffart2013yago2}, respectively.  
Specifically, DW-NB has more than 84,911 unique entities and contains 50,000 aligned entities, DY-NB has more than 58,858 unique entities and contains 15,000 aligned entities.
$36\%$ of the aligned entities have different entity names, which makes the datasets more realistic and the entity alignment task more challenging. The statistics of the datasets are summarized in Table \ref{table-dataset_statistics}. 

\subsection{Implementation Details}
\label{kba-experiments-hyperparameters}

We use grid search to find the best hyperparameters for TransAlign. We choose the embeddings dimensionality $d$ among $\{50, 75, 100, 200\}$, the learning rate of the Adam optimizer among $\{0.001, 0.01, 0.1\}$, and the margin $\gamma$ among $\{1, 5, 10\}$. We train TransAlign with a batch size of $100$ and a maximum of $400$ epochs.
We compare with representative state-of-the-art methods, and have used the hyper-parameters suggested by their corresponding papers.

\begin{table*}[htp]
	\centering
	\caption{The effect of the amount of seed entity alignments on EA accuracy in terms of Hits@k (\%). The numbers with bold/underline indicate the highest/sub-optimal values in each group compared to baseline methods.}
	\label{table-exp-1-results}
	\resizebox{0.9\textwidth}{!}{%
		\begin{tabular}{c|l|rr|rr|rr|rr|rr}
			\toprule[2pt]
			\midrule
			\multicolumn{2}{c|}{\multirow{2}[4]{*}{Method}} & \multicolumn{2}{c|}{Seed: 10\%} & \multicolumn{2}{c|}{Seed: 20\%} & \multicolumn{2}{c|}{Seed: 30\%} & \multicolumn{2}{c|}{Seed: 40\%} & \multicolumn{2}{c}{Seed: 50\%} \\
			\cmidrule{3-12}    \multicolumn{2}{c|}{} & \multicolumn{1}{l}{Hits@1} & \multicolumn{1}{l|}{Hits@10} & \multicolumn{1}{l}{Hits@1} & \multicolumn{1}{l|}{Hits@10} & \multicolumn{1}{l}{Hits@1} & \multicolumn{1}{l|}{Hits@10} & \multicolumn{1}{l}{Hits@1} & \multicolumn{1}{l|}{Hits@10} & \multicolumn{1}{l}{Hits@1} & \multicolumn{1}{l}{Hits@10} \\
			\midrule
			\midrule
			\multicolumn{12}{c}{DW-NB} \\
			\midrule
			\multirow{8}[-2]{*}{\begin{sideways}Translation-based\end{sideways}} & MTransE & 2.82  & 10.45 & 5.42  & 18.72 & 7.88  & 25.75 & 10.42 & 31.44 & 12.98 & 36.00 \\
			& IPTransE & 5.98  & 13.45 & 7.54  & 18.78 & 12.90 & 24.61 & 16.32 & 32.86 & 23.54 & 35.98 \\
			& BootEA & 8.12  & 16.15 & 12.54 & 20.13 & 17.92 & 28.38 & 21.46 & 35.16 & 25.44 & 37.57 \\
			& TransEdge & 22.98 & 48.12 & 38.29 & 56.22 & 45.27 & 68.95 & 49.26 & 75.25 & 54.85 & 79.68 \\
			& \underline{JAPE} & 4.62  & 7.87  & 8.62  & 14.43 & 12.57 & 19.96 & 17.20 & 27.32 & 19.91 & 30.63 \\
			& \underline{MultiKE} & 80.25 & 87.58 & 82.56 & 88.92 & 84.06 & 90.05 & 84.87 & 91.24 & 85.21 & 95.06 \\
			& \underline{AttrE} & \underline{87.98} & 95.80 & \underline{87.98} & 95.80 & \underline{87.98} & 95.80 & \underline{87.98} & 95.80 & 87.98 & 95.80 \\
			& \underline{TransAlign-W} & 87.81 & \underline{95.86} & 87.81 & \underline{95.86} & 87.81 & \underline{95.86} & 87.81 & \underline{95.86} & 87.81 & \underline{95.86} \\
			& \underline{TransAlign-A} & \textbf{88.73} & \textbf{96.91} & \textbf{88.73} & \textbf{96.91} & \textbf{88.73} & \textbf{96.91} & \textbf{88.73} & \textbf{96.91} & \textbf{88.73} & \textbf{96.91} \\
			\midrule
			\multirow{11}[2]{*}{\begin{sideways}GNN-based\end{sideways}} & MuGNN & 13.49 & 37.79 & 20.96 & 49.28 & 26.92 & 56.77 & 31.09 & 61.43 & 34.41 & 64.96 \\
			& AliNet & 14.58 & 31.46 & 18.55 & 35.84 & 24.34 & 50.46 & 28.39 & 55.46 & 35.31 & 58.22 \\
			& KECG  & 18.95 & 34.17 & 24.32 & 40.78 & 30.24 & 48.66 & 35.29 & 52.12 & 39.40 & 62.31 \\
			& \underline{GCN-Align} & 12.40 & 30.18 & 20.04 & 41.56 & 24.76 & 48.52 & 29.02 & 53.43 & 31.80 & 56.20 \\
			& \underline{HGCN} & 58.08 & 62.15 & 63.14 & 68.15 & 78.97 & 86.51 & 84.25 & 90.75 & 88.54 & 91.54 \\
			& \underline{GMNN} & 71.32 & 74.24 & 75.34 & 79.23 & 80.98 & 82.23 & 82.67 & 85.87 & 84.59 & 88.64 \\
			& \underline{RDGCN} & 59.22 & 62.98 & 64.22 & 68.98 & 79.02 & 87.12 & 85.34 & 90.45 & 88.21 & 93.23 \\
			& \underline{CEA}  & 50.13 & 52.31 & 63.25 & 64.12 & 80.32 & 84.21 & 84.34 & 85.54 & 86.58 & 88.34 \\
			& \underline{MRAEA} & 53.75 & 54.74 & 64.58 & 66.12 & 81.54 & 85.97 & 83.54 & 86.02 & 84.06 & 87.55 \\
			& \underline{NMN}  & 51.45 & 59.78 & 68.21 & 72.54 & 84.03 & 88.21 & 85.65 & 90.54 & \underline{88.69} & 95.46 \\
			\midrule
			\midrule
			\multicolumn{12}{c}{DY-NB} \\
			\midrule
			\multirow{8}[-2]{*}{\begin{sideways}Translation-based\end{sideways}} & MTransE & 0.01  & 0.15  & 0.01  & 0.39  & 0.08  & 0.68  & 0.08  & 1.39  & 0.13  & 1.89 \\
			& IPTransE & 1.54  & 9.87  & 5.67  & 25.76 & 14.55 & 36.45 & 15.77 & 45.81 & 17.33 & 52.18 \\
			& BootEA & 2.15  & 14.19 & 8.47  & 38.15 & 15.77 & 48.32 & 17.22 & 57.15 & 19.24 & 58.14 \\
			& TransEdge & 22.98 & 47.50 & 37.85 & 64.85 & 48.98 & 72.15 & 58.95 & 76.54 & 62.49 & 78.54 \\
			& \underline{JAPE} & 0.70  & 1.83  & 1.57  & 3.37  & 1.40  & 3.27  & 1.37  & 1.77  & 2.37  & 4.97 \\
			& \underline{MultiKE} & 81.87 & 88.05 & 82.11 & 89.26 & 84.97 & 90.84 & 87.22 & 92.05 & 89.25 & 93.58 \\
			& \underline{AttrE} & \underline{90.44} & 94.23 & \underline{90.44} & 94.23 & \underline{90.44} & 94.23 & \underline{90.44} & 94.23 & 90.44 & 94.23 \\
			& \underline{TransAlign-W} & 90.42 & \underline{94.35} & 90.42 & \underline{94.35} & 90.42 & \underline{94.35} & 90.42 & \underline{94.35} & 90.42 & 94.35 \\
			& \underline{TransAlign-A} & \textbf{91.27} & \textbf{95.62} & \textbf{91.27} & \textbf{95.62} & \textbf{91.27} & \textbf{95.62} & \textbf{91.27} & \textbf{95.62} & \textbf{91.27} & \textbf{95.62} \\
			\midrule
			\multirow{11}[2]{*}{\begin{sideways}GNN-based\end{sideways}} & MuGNN & 19.16 & 51.41 & 27.40 & 62.69 & 31.60 & 68.56 & 34.73 & 71.24 & 37.15 & 74.07 \\
			& AliNet & 13.54 & 28.53 & 14.25 & 31.69 & 25.39 & 58.31 & 28.98 & 56.12 & 34.59 & 64.12 \\
			& KECG  & 11.19 & 23.65 & 14.89 & 27.25 & 20.95 & 34.48 & 22.81 & 35.44 & 24.71 & 37.15 \\
			& \underline{GCN-Align} & 8.56  & 25.09 & 17.88 & 43.88 & 24.36 & 53.43 & 31.29 & 62.44 & 33.56 & 67.88 \\
			& \underline{HGCN} & 52.54 & 64.51 & 65.87 & 77.40 & 71.14 & 85.64 & 71.45 & 85.64 & 74.54 & 87.48 \\
			& \underline{GMNN} & 62.34 & 70.34 & 64.32 & 67.34 & 75.57 & 77.47 & 78.65 & 82.65 & 82.34 & 85.62 \\
			& \underline{RDGCN} & 53.13 & 65.30 & 67.28 & 78.21 & 74.54 & 85.22 & 77.45 & 87.43 & 78.67 & 89.45 \\
			& \underline{CEA}  & 55.24 & 58.97 & 64.35 & 65.42 & 74.56 & 78.42 & 77.78 & 80.95 & 78.91 & 83.24 \\
			& \underline{MRAEA} & 52.46 & 53.20 & 60.33 & 64.54 & 73.71 & 78.52 & 74.25 & 78.66 & 76.22 & 80.15 \\
			& \underline{NMN}  & 55.74 & 64.78 & 62.54 & 70.54 & 75.87 & 80.54 & 84.55 & 88.69 & \underline{90.78} & \underline{94.77} \\
			\midrule
			\midrule[2pt]
			\multicolumn{12}{l}{* Methods that use attribute triples are \ul{underlined}. The rest tables and figures follow this convention.}\\
			\multicolumn{12}{l}{* AttrE, TransAlign-W and TransAlign-A do not use any seed alignments.}
		\end{tabular}%
	}
	\vspace{-2mm}
\end{table*}%

\subsection{Compared Methods} 
\label{kba-experiments-models}
We propose TransAlign with two novel predicate alignment algorithms to achieve a fully automatic method for knowledge graph alignment. 
Here, we use \textbf{TransAlign-W} and \textbf{TransAlign-A} to represent TransAlign with weighted sum function and attention-based function respectively, which are the predicate embedding algorithms described in Section~\ref{kba-predicate-embedding}. 
Other compared existing entity alignment methods are described as follows.

	 \textbf{MTransE}~\cite{chen2017multilingual} is the state-of-the-art embedding-based alignment method built on top of TransE. MTransE learns a transition matrix from seed alignments to yield a unified embedding space from two KGs.
	 \textbf{IPTransE}~\cite{IPTransE2017} is an improved version of TransE. IPTransE adopts bootstrapping and has two soft strategies to add newly-aligned entities to the seeds to mitigate error propagation. 
	 \textbf{BootEA}~\cite{bordes2013translating} models EA as a one-to-one classification problem where the counterpart of an entity is regarded as the label of the entity. It iteratively learns the classifier via bootstrapping from both labeled and unlabeled data. 
	 \textbf{TransEdge}~\cite{sun2019transedge} proposes an edge-centric translational embedding method addressing the deficiency of TransE in that its relation predicate embeddings are entity-independent. 
	 \textbf{JAPE}~\cite{sun2017cross} is another state-of-the-art embedding-based entity alignment method built on top of TransE. It combines the relation triples with masked attribute triples. A masked attribute triple is an attribute triple in which its object is replaced by its data type.
	 \textbf{MultiKE}~\cite{MultiKE2019} uses multi-view learning on various kinds of features. The embedding module of MultiKE divides the features of KGs into three subsets called views: name view, relation view, and attribute view. Entity embeddings are learned for each view and then combined.
	 \textbf{AttrE}~\cite{trisedya2019entity} is the first method that makes use of attribute values and the only EA method that needs no seed alignments.
	 \textbf{MuGNN} \cite{cao2019multi} is the state-of-the-art embedding-based entity alignment  built on top of GCN, which uses two GCN as different channels to encode a KG. One channel is for completing missing links in a KG, and the other channel is for filtering unnecessary entities. Both channels are combined using a pooling layer. 
	 \textbf{AliNet}~\cite{sun2020knowledge} learns entity embeddings by a controlled aggregation of entity neighborhood information, and shares similar neighborhood structures by considering both direct and distant neighbors. 
	 \textbf{KECG}~\cite{KECG2019} aims to reconcile the issue of structural heterogeneity between KGs by jointly training both a GAT-based cross-graph module and a TransE-based knowledge embedding module.
	 \textbf{GCN-Align}~\cite{GCN-Align2018} is the first study on GNN-based EA, which learns entity embeddings from structural information of entities and exploits attribute triples by treating them as relation triples.
	 \textbf{HGCN}~\cite{HGCN2019} explicitly utilizes relation representation to improve the alignment process in EA. It incorporates the relation information by jointly learning entity and relation predicate embeddings.
	 \textbf{GMNN}~\cite{GMNN2019} formulates the EA task as graph matching between two topic entity graphs. It uses a graph matching module to model the similarity of two topic entity graphs, which indicates the probability of the two corresponding entities being aligned.
     \textbf{RDGCN}~\cite{RDGCN2019} utilizes relation information and extends GCNs with highway gates to capture the neighborhood structural information. It differs from HGCN in that it incorporates relation information by the attentive interaction.
     \textbf{CEA}~\cite{CEA2020} proposes a collective EA method which considers the dependency of alignment decisions among entities. It uses structural, semantic, and string signals to capture different aspects of the similarity between entities in the source and the target KGs, which are represented by three separate similarity matrices.
    \textbf{MRAEA}~\cite{MRAEA2020} considers meta relation semantics including relation predicates, relation direction, and inverse relation predicates, in addition to structural information learned from merely the structure of relation triples. The meta relation semantics are integrated into structural embedding via meta-relation-aware embedding and relation-aware GAT.
	\textbf{NMN}~\cite{NMN2020} aims to tackle the structural heterogeneity between KGs. The method learns both the KG structure information and the neighborhood difference so that the similarities between entities can be better captured in the presence of structural heterogeneity.
\vspace{-3mm}

\subsection{Entity Alignment Results} 
\label{kba-entity-alignment-results}

This experiment evaluates the accuracy of EA while varying the amount of seed entity alignments used for training from $10\%$ to $50\%$ of the total available set of seed entity alignments (50,000 for DW-NB and 7,500 for DY-NB). We evaluate the performance of TransAlign (note that it does not need any seed alignments) using $\mathbf{Hits@k} (k=1,10)$ (i.e., the proportion of correctly aligned entities ranked in the top $k$ predictions). A higher value indicates better performance. 

Table~\ref{table-exp-1-results} shows the results on the DWY-NB benchmark datasets \cite{zhang2022benchmark}. Some of the results of the compared methods are obtained from \cite{zhang2022benchmark}. We found that two variations of TransAlign, TransAlign-W and TransAlign-A, are significantly better than all the other methods. 
The underlined methods from both translation- and GNN-based methods exploit attribute triples, and we can see that on average, the methods that exploit attribute triples achieve much better performance than the methods that do not. 

To better investigate the effect of seed alignments on existing methods, we gradually increase the proportion of seed alignments, from $10\%$ to $50\%$. The results show that the accuracy of these methods are increasing along with the portion of seed alignment. 
When fewer seed entity alignments are available, TransAlign, AttrE and MultiKE have much better accuracy than the others. This is because they make better use of various types of features such as attributes and relation predicates. The accuracy of AttrE and TransAlign do not change while varying the amount of seeds since they do not use seed alignments, so when seed alignments are hardly available, they are the clear winners.
In particular, the predicate alignments of TransAlign are learned from the predicate-proximity-graph, while the entity alignments are learned from the combination of the structure embeddings and the attribute embeddings.
The accuracy of TransAlign-A is better than TransAlign-W, which shows the importance to capture both the distinctive and noisy entity types, as done by TransAlign-A. 

\begin{figure*}[htp]
    \centering
	\begin{tikzpicture}
	\begin{groupplot}[
	width = 0.72\textwidth,
	height = 0.2\textwidth,
	legend pos= outer north east 
	]
	
	\nextgroupplot[
	enlarge x limits=0.08,
	axis x line*=bottom,
	axis y line*=left,
	ybar,
	ylabel=hits@1,
	ymin=0,
	ymax=100,
	xtick={TransAlign-W, TransAlign-W*,  TransAlign-A, TransAlign-A*,   MultiKE}, 
	symbolic x coords={{TransAlign-W}, {TransAlign-W*},  {TransAlign-A}, {TransAlign-A*},  {MultiKE}},
	bar width=8pt,
	]
	\addplot[pattern=north east lines] coordinates { (TransAlign-W, 87.81) (TransAlign-W*, 12.58) (TransAlign-A, 88.73) (TransAlign-A*, 12.88)  (MultiKE, 84.06) };
	\addplot[color=black, fill=black!25, postaction={pattern=grid}] coordinates {(TransAlign-W, 90.42) (TransAlign-W*, 14.55) (TransAlign-A, 91.27) (TransAlign-A*, 15.21)  (MultiKE, 84.97) };
	
	\addlegendimage{pattern=north east lines}
	\addlegendentry{ DW-NB }
	\addlegendimage{color=black, fill=black!25, postaction={pattern=grid}}
	\addlegendentry{ DY-NB }
	
	\end{groupplot}
	\end{tikzpicture}
	\vspace{-2mm}
	\caption{The effect of attribute embedding module.}
	\label{fig-exp-2-results}
\end{figure*}
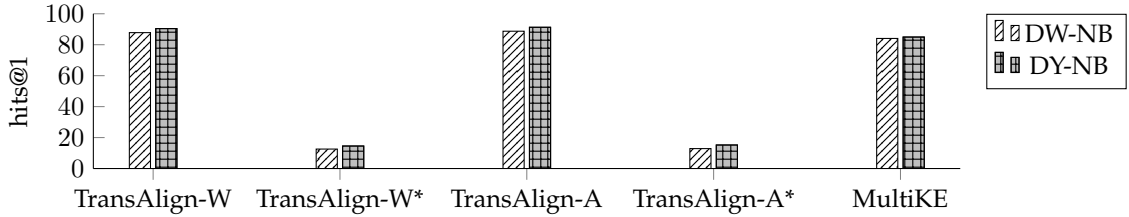

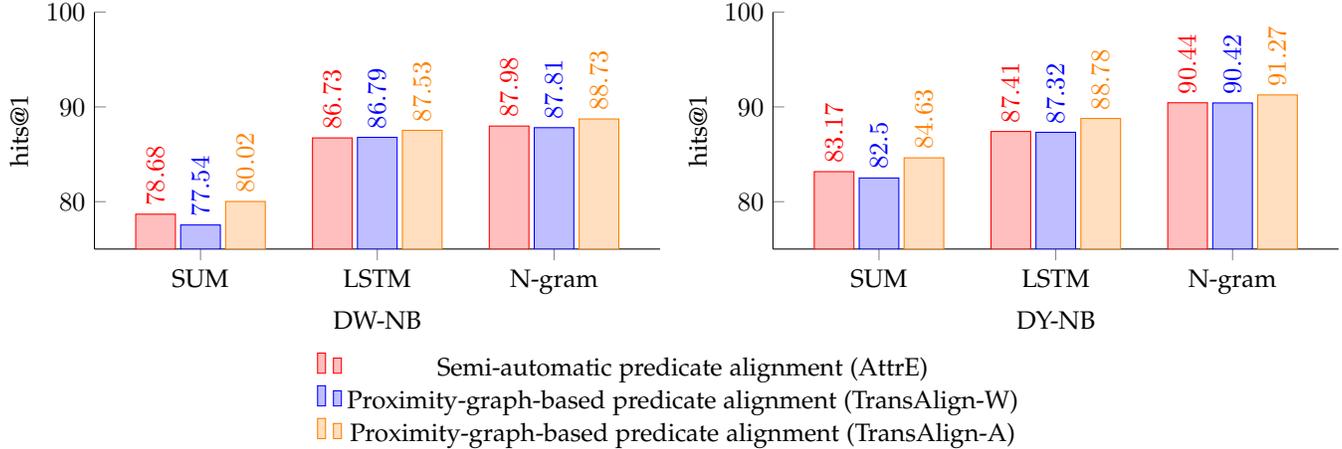
\begin{figure*}
	\begin{tikzpicture}
	\begin{groupplot}[
	group style = {group size = 2 by 1, horizontal sep=1.5cm},
	width = 0.5\textwidth,
	height = 0.26\textwidth,
	]
	
	
	\nextgroupplot[
	enlarge x limits=0.3,
	axis x line*=bottom,
	axis y line*=left,
	ybar,
	ylabel=hits@1,
	ymin=75,
	ymax=100,
	xlabel=DW-NB,
	xtick={SUM, LSTM, N-gram},
	symbolic x coords={{SUM},{LSTM},{N-gram}},
	bar width=15pt,
	nodes near coords,
	every node near coord/.append style={rotate=90, anchor=west},
	]
	\addplot[color=red, fill=red!25] coordinates {(SUM,78.68) (LSTM,86.73) (N-gram,87.98)};
	\addplot[color=blue, fill=blue!25] coordinates {(SUM,77.54) (LSTM,86.79) (N-gram,87.81)};
	\addplot[color=orange, fill=orange!25] coordinates {(SUM,80.02) (LSTM,87.53) (N-gram,88.73)};
	
	\nextgroupplot[
	enlarge x limits=0.3,
	axis x line*=bottom,
	axis y line*=left,
	ybar,
	ylabel=hits@1,
	ymin=75,
	ymax=100,
	xlabel=DY-NB,
	xtick={SUM, LSTM, N-gram},
	symbolic x coords={{SUM},{LSTM},{N-gram}},
	bar width=15pt,
	nodes near coords,
	every node near coord/.append style={rotate=90, anchor=west},
	legend style={at={($(3,3)+(2cm,2cm)$)},legend columns=1,fill=none,draw=none,anchor=center,align=left},
	legend to name=lg
	]
	\coordinate (c2) at (rel axis cs:1,1);
	\addplot[color=red, fill=red!25] coordinates {(SUM,83.17) (LSTM,87.41) (N-gram,90.44)};
	\addplot[color=blue, fill=blue!25] coordinates {(SUM,82.5) (LSTM,87.32) (N-gram,90.42)};
	\addplot[color=orange, fill=orange!25] coordinates {(SUM,84.63) (LSTM,88.78) (N-gram,91.27)};
	\addlegendimage{red!25}
	\addlegendentry{Semi-automatic predicate alignment (AttrE)}
	\addlegendimage{blue!25}
	\addlegendentry{Proximity-graph-based predicate alignment (TransAlign-W)}
	\addlegendimage{blue!25}
	\addlegendentry{Proximity-graph-based predicate alignment (TransAlign-A)}
	\end{groupplot}
	\node[below] at (current bounding box.south){\pgfplotslegendfromname{lg}};
	\end{tikzpicture}
	\vspace{-2mm}
	\caption{The effect of predicate embedding module.}
	\label{fig-kba-predicate-alignment}
\end{figure*}

\subsection{Ablation Study}
To investigate the effectiveness of the proposed modules in TransAlign, we conduct ablation test from two perspectives.
\subsubsection{Effect of Attribute Embedding Module} 
\label{kba-discussion-attribute}
To evaluate the effect of using attribute triples, we create a version of TransAlign-W that does not use attribute triples to compute the entity embeddings, i.e., it only uses relation triples; we call this version TransAlign-W*. Similarly, we create a version of TransAlign-A that does not use attribute triples, which we call TransAlign-A*. Figure~\ref{fig-exp-2-results} shows the accuracy of the four versions of TransAlign on the benchmark. We can see that the accuracy of TransAlign-W and TransAlign-A are much higher than that of TransAlign-W* and TransAlign-A*, respectively. This shows that our idea of using attribute triples 
is highly effective. We also put the accuracy of MultiKE in the figure for comparison since MultiKE is the most accurate one among other existing methods; the proportion of seed entity alignments used for MultiKE is $30\%$. TransAlign-W and TransAlign-A both outperform MultiKE.


We also show the effect of different attribute embedding algorithms in Fig.~\ref{fig-kba-predicate-alignment}. Here SUM, LSTM, and N-gram denote three algorithms with different attribute embedding functions, as described in Section \ref{kba-attribute-character-embedding}. 
We see that the N-gram compositional function gives the best performance. This is because the N-gram compositional function better preserves string similarity when mapping attribute strings to their vector representations than the other two functions. 

\subsubsection{Effect of Predicate Embedding Module} 
\label{kba-discussion-predicate}

To evaluate the effect of predicate embedding module proposed in Section~\ref{kba-predicate-embedding}, we compare with the semi-automatic predicate alignment module in our previously proposed method AttrE \cite{trisedya2019entity}. 
From Fig.~\ref{fig-kba-predicate-alignment}, we see that the predicate embedding module helps our entity alignment method achieve comparable performance in terms of $hits@1$. 

The same predicate may be stored in different surface forms in the KGs, e.g., one KG has the attribute predicate \texttt{birth\_date} while the other KG has the attribute predicate \texttt{date\_of\_birth}. 
Previous methods exploit seed attribute predicate alignments and seed attribute alignments to address this difference. 
AttrE first exploits attribute triples without seed attribute alignments to learn a unified attribute vector space in the semi-automatic procedure, which requires manual intervention. In comparison, both our newly proposed TransAlign-W and TransAlign-A do not need manual intervention, while they both yield  competitive results. In particular, TransAlign-A achieves the best performance since it enriches the related entity types information via attention mechanism. 
\vspace{-2mm}

%
%
\begin{table}[hbt!]
	\begin{center}
		\caption{The effect on downstream link prediction task in terms of Hits@10 (\%). The numbers with bold/underline indicate the highest/sub-optimal values in each group compared to baseline methods.}
		\label{table-exp-4-results}
		\resizebox{\columnwidth}{!}{%
    		\begin{tabular}{@{}lllllll}
                \toprule[2pt]
                \midrule
                \multicolumn{1}{@{}c|}{\multirow{2}[4]{*}{Method}} & \multicolumn{3}{c|}{DW-NB (seed)} & \multicolumn{3}{c}{DY-NB (seed)} \\
            \cmidrule{2-7}    \multicolumn{1}{c|}{} & \multicolumn{1}{c}{(10\%)} & \multicolumn{1}{c}{(30\%)} & \multicolumn{1}{c|}{(50\%)} & \multicolumn{1}{c}{(10\%)} & \multicolumn{1}{c}{(30\%)} & \multicolumn{1}{c}{(50\%)} \\
                \midrule[1.5pt]
                 \multicolumn{1}{l|}{\ul{TransAlign-A}} & \multicolumn{1}{r}{\textbf{88.93}} & \multicolumn{1}{r}{\underline{88.93}} & \multicolumn{1}{r|}{\underline{88.93}} & \multicolumn{1}{r}{\textbf{98.82}} & \multicolumn{1}{r}{\underline{98.82}} & \multicolumn{1}{r}{\textbf{98.82}} \\
                \multicolumn{1}{l|}{\ul{MultiKE}} & \multicolumn{1}{r}{\underline{88.76}} & \multicolumn{1}{r}{\textbf{88.98}} & \multicolumn{1}{r|}{\textbf{89.52}} & \multicolumn{1}{r}{98.62} & \multicolumn{1}{r}{\textbf{98.87}} & \multicolumn{1}{r}{98.07} \\
                \multicolumn{1}{l|}{\ul{AttrE}} & \multicolumn{1}{r}{88.50} & \multicolumn{1}{r}{88.50} & \multicolumn{1}{r|}{88.50} & \multicolumn{1}{r}{\underline{98.75}} & \multicolumn{1}{r}{98.75} & \multicolumn{1}{r}{\underline{98.75}} \\
                \multicolumn{1}{l|}{\ul{TransAlign-W}} & \multicolumn{1}{r}{88.41} & \multicolumn{1}{r}{88.41} & \multicolumn{1}{r|}{88.41} & \multicolumn{1}{r}{98.66} & \multicolumn{1}{r}{98.66} & \multicolumn{1}{r}{98.66} \\
                \multicolumn{1}{l|}{TransE} & \multicolumn{1}{r}{87.45} & \multicolumn{1}{r}{87.45} & \multicolumn{1}{r|}{87.45} & \multicolumn{1}{r}{98.42} & \multicolumn{1}{r}{98.42} & \multicolumn{1}{r}{98.42} \\
                \multicolumn{1}{l|}{TransEdge} & \multicolumn{1}{r}{85.27} & \multicolumn{1}{r}{85.71} & \multicolumn{1}{r|}{86.40} & \multicolumn{1}{r}{93.24} & \multicolumn{1}{r}{93.54} & \multicolumn{1}{r}{93.76} \\
                \multicolumn{1}{l|}{\ul{JAPE}} & \multicolumn{1}{r}{83.24} & \multicolumn{1}{r}{83.71} & \multicolumn{1}{r|}{83.09} & \multicolumn{1}{r}{75.03} & \multicolumn{1}{r}{75.32} & \multicolumn{1}{r}{75.66} \\
                \multicolumn{1}{l|}{IPTransE} & \multicolumn{1}{r}{81.06} & \multicolumn{1}{r}{81.23} & \multicolumn{1}{r|}{81.78} & \multicolumn{1}{r}{93.10} & \multicolumn{1}{r}{93.55} & \multicolumn{1}{r}{93.91} \\
                \multicolumn{1}{l|}{BootEA} & \multicolumn{1}{r}{80.41} & \multicolumn{1}{r}{80.90} & \multicolumn{1}{r|}{81.66} & \multicolumn{1}{r}{94.11} & \multicolumn{1}{r}{94.54} & \multicolumn{1}{r}{94.85} \\
                \multicolumn{1}{l|}{MTransE} & \multicolumn{1}{r}{80.10} & \multicolumn{1}{r}{80.33} & \multicolumn{1}{r|}{80.69} & \multicolumn{1}{r}{93.81} & \multicolumn{1}{r}{94.31} & \multicolumn{1}{r}{94.74} \\
                \midrule
                \midrule[2pt]
                \\
            \end{tabular}%
		}
	\end{center}
	\vspace{-5mm}
\end{table}
\subsection{Effect of the Alignment Method on KG embeddings} 
\label{kba-discussion2}

We further evaluate the effect of TransAlign on KG embeddings. Since the training in EA methods optimize for two objectives, i.e.,  \textit{KG embeddings} and \textit{alignment of two KGs} (either jointly or alternatively), rather than merely KG embeddings, so it might not produce the best KG embeddings. 
This experiment evaluates how the quality of the KG embeddings obtained from EA methods are affected compared to the KG embeddings obtained from pure KG embedding methods (TransE for translation-based and GCN for GNN-based methods) via downstream applications of KGs. 
Following previous studies in EA methods \cite{sun2020benchmarking,zhao2020experimental}, we conduct experiments using a common downstream task \emph{link prediction} for this purpose, detailed as follows. 
The link prediction task aims to predict $t$ given $h$ and $r$ of a relation triple. 
Specifically, first a relation triple is corrupted by replacing its tail entity with all the entities in the dataset. Then, the corrupted triples are ranked in ascending order by the plausibility score computed as $\boldsymbol{h} + \boldsymbol{r} - \boldsymbol{t}$. Since true triples (i.e., the triples in a KG) are expected to have smaller plausibility scores and rank higher in the list than the corrupted ones, hits@10 (whether the true triples are in the top-10) is used as the measure for the link prediction task.

Table~\ref{table-exp-4-results} shows the accuracy of link prediction on DW-NB and DY-NB with $10\%$, $30\%$, and $50\%$ of seed entity alignments. The accuracy increases with the amount of seed alignments but not significantly.
As mentioned earlier, the KG embeddings obtained from the KG alignment methods may not be optimized for downstream tasks. However, TransAlign-A still achieves high accuracy, always in top-2 and top-1 in half of the cases, which shows that the learned predicate embeddings can also project entities into a unified embedding space. 

\vspace{-3mm}
\section{Conclusion and Future Work} 
\label{kba-conclusion}

We have presented TransAlign -- the first fully automatic method for KG alignment. To achieve that, we proposed attribute character embeddings and predicate-proximity-graph embeddings to compute a unified vector space for the entity and predicate embeddings from two KGs. Experimental results show that TransAlign outperforms the competitors consistently.
Further results on knowledge base completion show that our joint learning of the entity, predicate, and attribute embeddings can capture the similarity between entities and predicates both within a KG and across KGs. 

TransAlign is built on top of a translation-based KG embedding method, e.g., TransE. For future work, we plan to investigate the potential of exploiting our ideas via other KG embedding methods, such as GNN-based and graph-transformer-based ones. 

\ifCLASSOPTIONcaptionsoff
  \newpage
\fi
\vspace{-3mm}
\bibliographystyle{IEEEtran}
\bibliography{bib/Trisedya_TKDE_AAAI19}
\vspace{-1cm}
\begin{IEEEbiography}[{\includegraphics[width=1in,height=1.25in,clip,keepaspectratio]{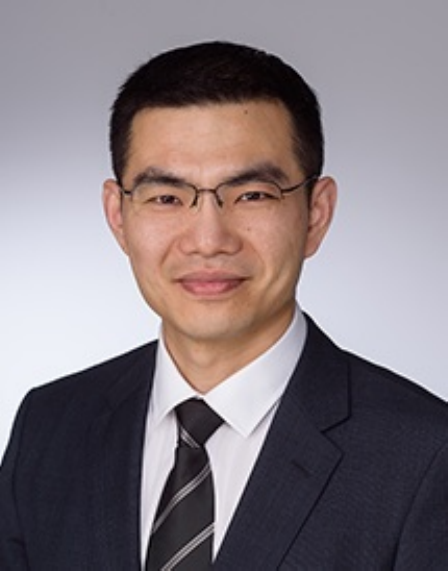}}]{Rui Zhang}
is a visiting Professor at Tsinghua University. His research interests include big data, data mining, and machine learning. Professor Zhang has won several awards, including Future Fellowship by the Australian Research Council in 2012, Chris Wallace Award for Outstanding Research by the Computing Research and Education Association of Australasia in 2015, and Google Faculty Research Award in 2017.
\end{IEEEbiography}
\vspace{-1cm}
\begin{IEEEbiography}[{\includegraphics[width=1.0in,height=1.0in,clip,keepaspectratio]{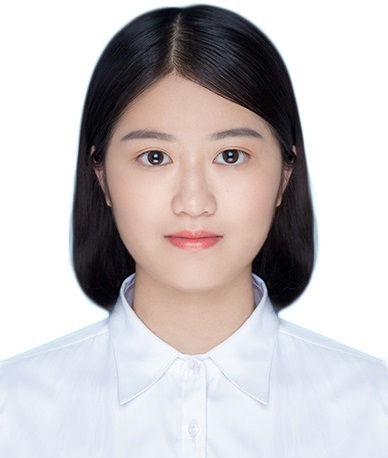}}] 
{Xiaoyan Zhao} is currently a PhD student with the Department of Systems Engineering and Engineering Management, the Chinese University of Hong Kong. She received the B.S. degree from Wuhan University of Technology in 2019, and M.E. degree from University of Chinese Academy of Sciences in 2022. Her research interests include natural language processing and knowledge graph.
\end{IEEEbiography}
\vspace{-1cm}
\begin{IEEEbiography}[{\includegraphics[width=1in,height=1.25in,clip,keepaspectratio]{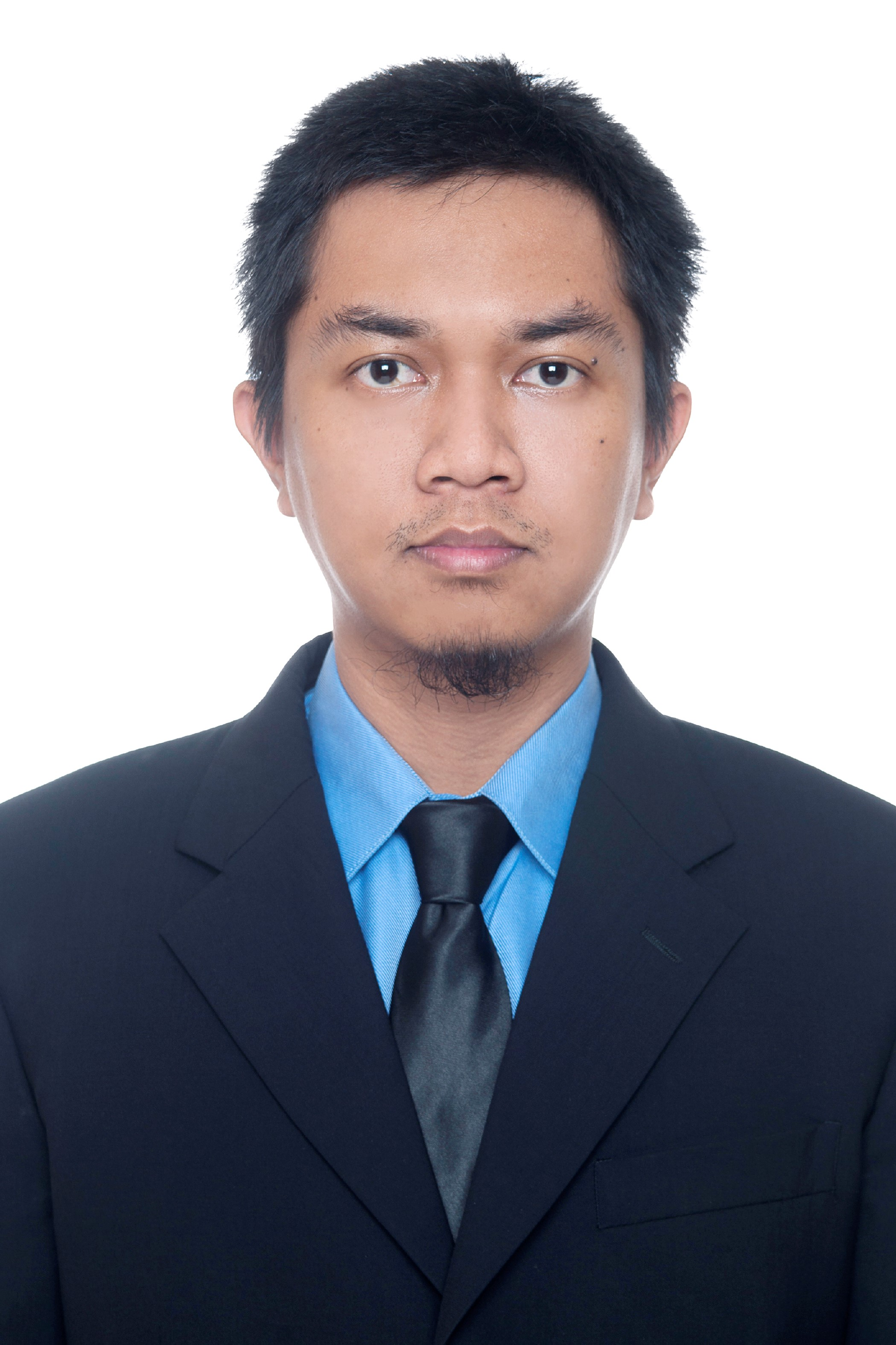}}]{Bayu Distiawan Trisedya}
is a Lecturer in the Faculty of Computer Science Universitas Indonesia. He was a Postdoctoral Research Fellow in the School of Computing and Information Systems at The University of Melbourne and RMIT University. He received his bachelor’s and Master’s degrees from Universitas Indonesia in 2009 and 2011, respectively. He received his PhD degree from The University of Melbourne in 2021. His research interest is knowledge graphs and natural language processing.
\end{IEEEbiography}
\vspace{-1cm}
\begin{IEEEbiography}[{\includegraphics[width=1.0in,height=1.0in,clip,keepaspectratio]{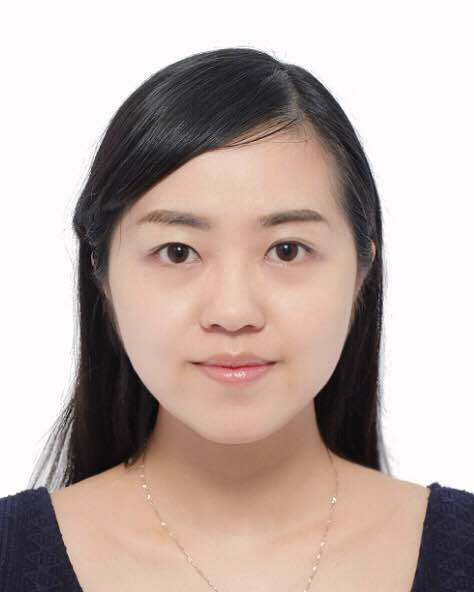}}]
{Min Yang} is currently an Associate Professor with the Shenzhen Institutes of Advanced Technology, Chinese Academy of Science. She received her Ph.D. degree from the University of Hong Kong in February 2017. Prior to that, she received her B.S. degree from Sichuan University in 2012.  Her current research interests include machine learning, deep learning and natural language processing.
\end{IEEEbiography}
\vspace{-1cm}
\begin{IEEEbiography}[{\includegraphics[width=1.0in,height=1.0in,clip,keepaspectratio]{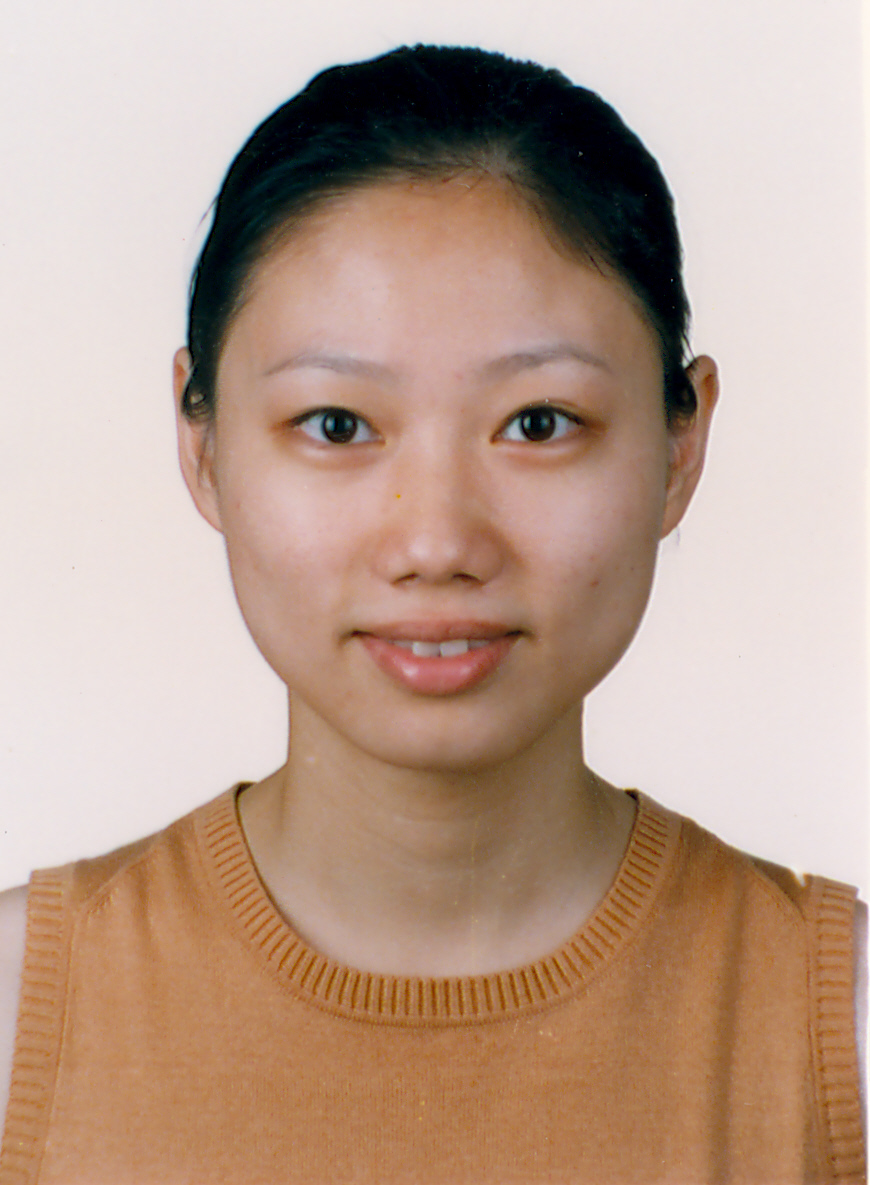}}]
{Hong Cheng} is a Professor in the Department of Systems Engineering and Engineering Management at the Chinese University of Hong Kong. She received her Ph.D. degree from University of Illinois at Urbana-Champaign in 2008. Her research interests include data mining, database systems, and machine learning. She received research paper awards at ICDE’07, SIGKDD’06 and SIGKDD’05, and the certificate of recognition for the 2009 SIGKDD Doctoral Dissertation Award. She is a recipient of the 2010 Vice-Chancellor’s Exemplary Teaching Award at the Chinese University of Hong Kong.
\end{IEEEbiography}
\vspace{-1cm}
\begin{IEEEbiography}[{\includegraphics[width=1in,height=1.25in,clip,keepaspectratio]{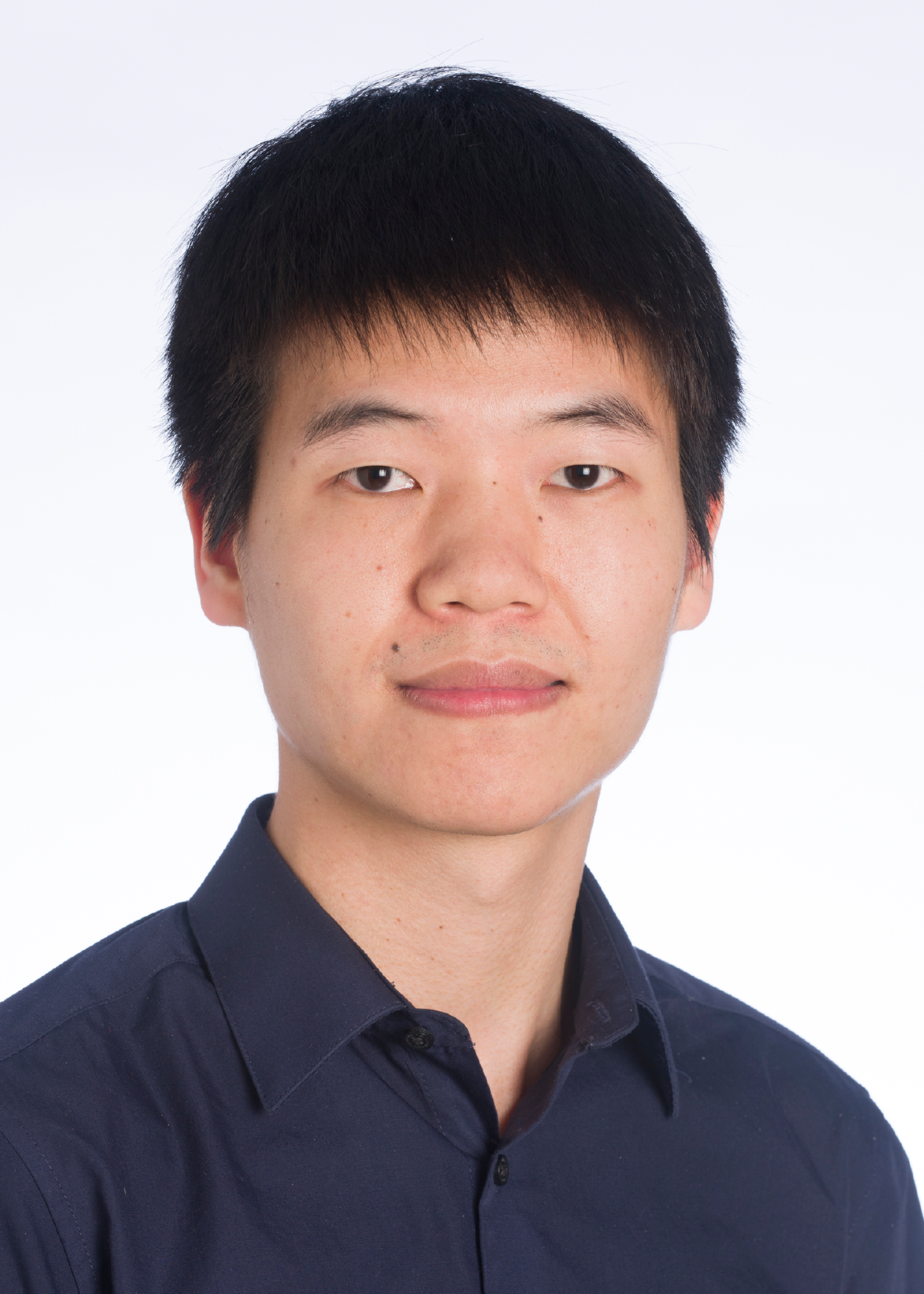}}]{Jianzhong Qi}
is a Senior Lecturer in the School of Computing and Information Systems at The University of Melbourne. He received his Ph.D. degree from The University of Melbourne in 2014. His research interests include machine learning and data management and analytics, with a focus on spatial, temporal, and textual data.
\end{IEEEbiography}
\vfill

\end{document}